\documentclass{article}
\usepackage[a4paper,margin=0.5in]{geometry}

\usepackage[utf8]{inputenc}
\usepackage[T1]{fontenc}
\usepackage{lineno}
\usepackage{float}
\usepackage{tabularx}
\usepackage{hyperref}
\usepackage[normalem]{ulem}
\usepackage{subfigure}
\usepackage{multirow}
\usepackage{multicol}
\usepackage{adjustbox}
\usepackage[mathscr]{euscript}
\def\BibTeX{{\rm B\kern-.05em{\sc i\kern-.025em b}\kern-.08em
    T\kern-.1667em\lower.7ex\hbox{E}\kern-.125emX}}
\usepackage{authblk} 
\usepackage{xcolor}
\usepackage{amsmath}
\usepackage{amssymb}

\usepackage[numbers]{natbib}

\title{RoboMNIST: A Multimodal Dataset for Multi-Robot Activity Recognition Using WiFi Sensing, Video, and Audio}

\author[1]{Kian Behzad}
\author[1]{Rojin Zandi}
\author[1]{Elaheh Motamedi}
\author[2]{Hojjat Salehinejad}
\author[1]{Milad Siami}

\affil[1]{Department of Electrical \& Computer Engineering, Northeastern University, Boston, MA, USA \\
\texttt{\{behzad.k, zandi.r, motamedi.e, m.siami\}@northeastern.edu}}
\affil[2]{Kern Center for the Science of Health Care Delivery and Department of Artificial Intelligence and Informatics, Mayo Clinic, Rochester, MN, USA \\
\texttt{salehinejad.hojjat@mayo.edu}}
\affil[*]{Corresponding author: \href{mailto:m.siami@northeastern.edu}{m.siami@northeastern.edu}}

\date{}

\begin{document}
\flushbottom
\maketitle

\begin{abstract}
We introduce a novel dataset for multi-robot activity recognition (MRAR) using two robotic arms integrating WiFi channel state information (CSI), video, and audio data. This multimodal dataset utilizes signals of opportunity, leveraging existing WiFi infrastructure to provide detailed indoor environmental sensing without additional sensor deployment. Data were collected using two Franka Emika robotic arms, complemented by three cameras, three WiFi sniffers to collect CSI, and three microphones capturing distinct yet complementary audio data streams. The combination of CSI, visual, and auditory data can enhance robustness and accuracy in MRAR. This comprehensive dataset enables a holistic understanding of robotic environments, facilitating advanced autonomous operations that mimic human-like perception and interaction. By repurposing ubiquitous WiFi signals for environmental sensing, this dataset offers significant potential aiming to advance robotic perception and autonomous systems. It provides a valuable resource for developing sophisticated decision-making and adaptive capabilities in dynamic environments.
\end{abstract}

\section*{Background \& Summary}
Signals of opportunity refer to the use of pre-existing, non-dedicated signals in the environment for secondary purposes beyond their original intent. WiFi signals, for instance, are primarily used for communication, but can also provide valuable information about the environment through channel state information (CSI). This data captures intricate details about the propagation of WiFi signals, including reflections, scattering, and absorption caused by objects and activities in the environment. By repurposing these ubiquitous WiFi signals, we can achieve comprehensive indoor environmental sensing without the need for additional sensor infrastructure.

Combining WiFi CSI with video and audio data creates a powerful multimodal system that significantly enhances activity recognition capabilities. Video data provides rich visual information, capturing spatial and temporal changes in the environment. Audio data adds another layer of contextual information, identifying auditory cues that correlate with specific activities. Integrating these modalities with CSI data leads to a more holistic understanding of the environment, enabling more accurate and reliable multi-robot activity recognition (MRAR). A true potential of such integration lies in the broader concept of sensor fusion. Sensor fusion combines data from multiple sensors to achieve a more accurate and comprehensive understanding of an environment than could be obtained from any single sensor. This approach not only improves accuracy, but also enhances system resilience and supports robust decision-making across various applications, including robotics and autonomous systems. For example, in scenarios where visual information is compromised due to obstructions or lack of line-of-sight, CSI data can detect movements and objects without requiring direct visual contact, compensating for the limitations of video and audio sensors. Such multimodal sensor fusion is critical for complex and dynamic environments, leveraging the complementary strengths of different sensing modalities. The effectiveness of this approach has been demonstrated in various applications, such as multi-sensor fusion for autonomous vehicle tracking, which integrates diverse sensor inputs to enhance object detection and tracking performance \cite{vinoth2024multi, celik2023wearable}.

The use of signals of opportunity, such as WiFi CSI, offers several key advantages. Firstly, it leverages existing infrastructure, reducing the cost and complexity associated with deploying additional sensors. This makes it an economically viable option for large-scale implementations. Secondly, the multimodal nature of the sensing network enhances robustness by compensating for the limitations of individual sensors. For instance, in scenarios where visual data may be obscured or audio data may be noisy, the complementary information from CSI can help maintain accurate activity recognition.

In recent decades, the integration of robotics and automated systems into various sectors has markedly transformed human life and industrial processes \cite{moran2007evolution}. The advent of multi-agent systems, where multiple robots collaborate, has subtly yet significantly enhanced task efficiency and adaptability \cite{prajapat2022near,hosseini2023Agent}. In healthcare, robots perform complex surgeries with unprecedented precision, reducing recovery times and improving outcomes. In manufacturing, automated assembly lines and robotic arms ensure consistent quality and high productivity, revolutionizing production techniques. Additionally, robots are deployed in hazardous environments, like disaster sites \cite{jorge2019survey}, minimizing human risks. 

In the realm of robotics, the integration of multimodal learning systems represents a significant leap towards achieving autonomous operations that closely mimic human-like perception and interaction with the environment \cite{duan2022multimodal}. This paper explores the application of such an advanced learning paradigm through the deployment of two Franka Emika robotic arms, a choice inspired by their versatility and precision in complex tasks \cite{haddadin2022franka}. To enrich the sensory framework essential for multimodal learning, we employ a comprehensive array of sensors: three cameras, three WiFi sniffers, and three microphones. Each modality is strategically chosen to capture distinct yet complementary data streams—visual, wireless signal-based, and auditory—thereby enabling a more holistic understanding of the robotic arms' surroundings and activities. This multi-sensory approach not only facilitates the robust perception required for intricate manipulations and interactions but also paves the way for groundbreaking advancements in autonomous robotic systems capable of sophisticated decision-making and adaptation in dynamic environments.

The MNIST dataset \cite{lecun1998gradient}, featuring handwritten digits classified into ten categories, was first introduced by LeCun et al. in 1998. At that time, the significant advancements and performance of deep learning techniques were unimaginable. Despite the current extensive capabilities of deep learning, the simple MNIST dataset remains the most widely used benchmark in the field, even surpassing CIFAR-$10$ \cite{krizhevsky2009learning} and ImageNet \cite{deng2009imagenet} in popularity according to Google Trends. Its simplicity hasn't diminished its usage, despite some in the deep learning community advocating for its decline \cite{xiao2017fashion}.

In this paper, we introduce the RoboMNIST dataset, an innovative extension of the traditional MNIST dataset tailored for robotic applications. This dataset features two Franka Emika robots writing different digits on an imaginary plane within a $3$D environment. Our sensor-rich modules, comprising CSI, video, and audio, capture comprehensive data from the environment. We have validated the dataset's integrity across individual data modalities through a series of experiments.

\section*{Methods}
\label{sec:methods}
\begin{figure}[htbp]
    \centering
    \includegraphics[width=0.8 \textwidth]{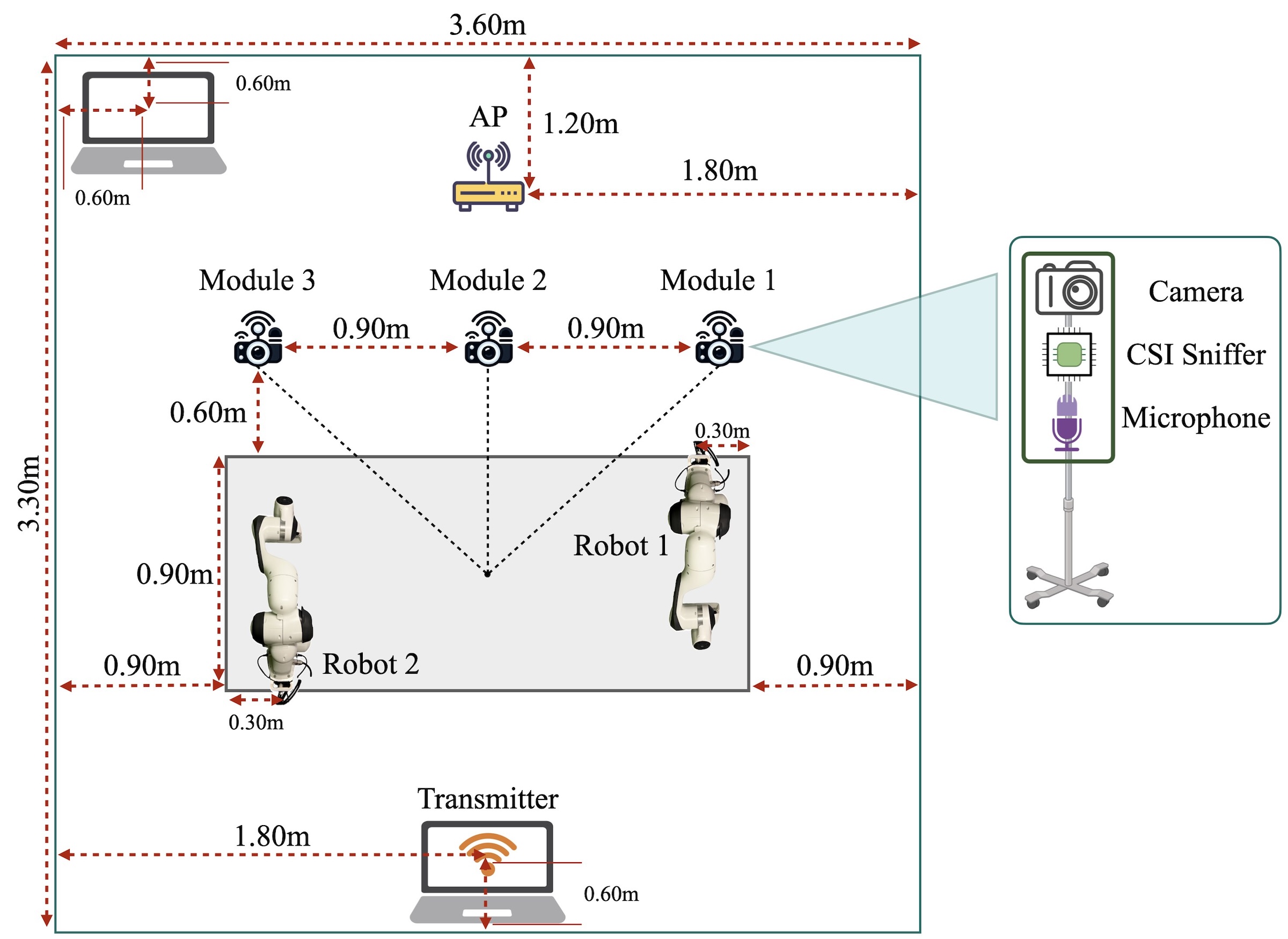}
    \caption{Floor plan of the data collection environment, where the robots are performing different activities while various sensors mounted on our sensor-rich modules capture data.}
    \label{fig:floor_plan}
\end{figure}

The data collection process was conducted in a laboratory, featuring desks, chairs, monitors, and various other office objects in the environment. The layout of the laboratory, along with its physical dimensions, is illustrated in Figure~\ref{fig:floor_plan}.

\subsection*{Hardware Specifications \& Communication}
\label{sec:methods:hardware_communication}

\begin{figure}[htbp]
    \centering
    \includegraphics[width=0.4 \textwidth]{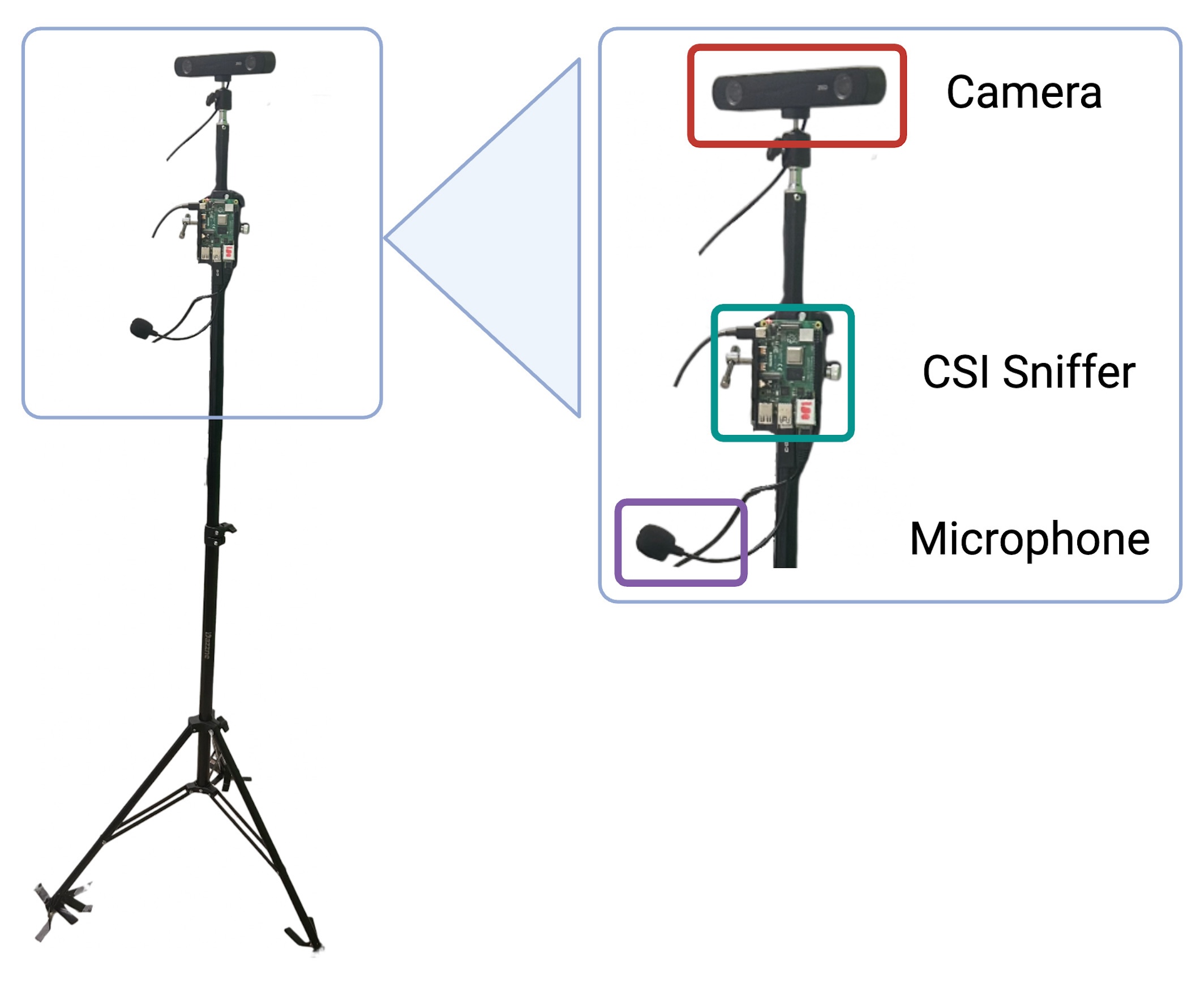}
    \caption{Sensor-rich module used in the data collection setup consisting of a Raspberry Pi to capture CSI, a stereo camera to capture video, and a microphone to capture audio.}
    \label{fig:sensor_rich_module}
\end{figure}

In all the experiments we have used three sensor-rich modules that are positioned in the room capturing data while the two robotic arms are performing different activities. Each sensor-rich module is capable of simultaneously capturing three modalities namely CSI, video, and audio from the environment. Each module is equipped with the following hardware:
\begin{itemize}
    \item \textbf{CSI}: A Raspberry Pi 4 Model B, integrated with the Nexmon project \cite{nexmon:project}, which passively captures the CSI data.
     \item \textbf{Video}: A ZED 2 Stereo Camera for video recording.
    \item \textbf{Audio}: A CG CHANGEEK Mini USB Microphone, featuring omni-directional directivity, to record audio.
\end{itemize}
Figure~\ref{fig:sensor_rich_module} depicts the sensor-rich modules used in our data collection setup. We use $M \in \{1, 2, 3\}$ to denote the modules based on the numbering notation in Figure~\ref{fig:floor_plan} in the rest of the paper.
To facilitate the collection of CSI data, an Apple Mac Mini equipped with the 802.11ax WiFi 6 standard served as the WiFi transmitter.

\begin{figure}[htbp]
\centering
\begin{minipage}{0.45\textwidth}
\centering
\begin{tabular}{|c|c|c|c|c|}
\hline
Frame & $a$ (m) & $d$ (m) & $\alpha$ (radian) & $\theta$ (radian) \\ \hline
Joint 1 & 0 & 0.333 & 0 & $q_1$ \\ \hline
Joint 2 & 0 & 0 & $-\pi/2$ & $q_2$ \\ \hline
Joint 3 & 0 & 0.316 & $\pi/2$ & $q_3$ \\ \hline
Joint 4 & 0.0825 & 0 & $\pi/2$ & $q_4$ \\ \hline
Joint 5 & -0.0825 & 0.384 & $-\pi/2$ & $q_5$ \\ \hline
Joint 6 & 0 & 0 & $\pi/2$ & $q_6$ \\ \hline
Joint 7 & 0.088 & 0 & $\pi/2$ & $q_7$ \\ \hline
Flange & 0 & 0.107 & 0 & 0 \\ \hline
End effector (EE) & 0 & 0.1034 & 0 & $\pi/4$ \\ \hline
\end{tabular}
\end{minipage}
\hfill
\begin{minipage}{0.45\textwidth}
\centering
\includegraphics[width=0.5 \textwidth]{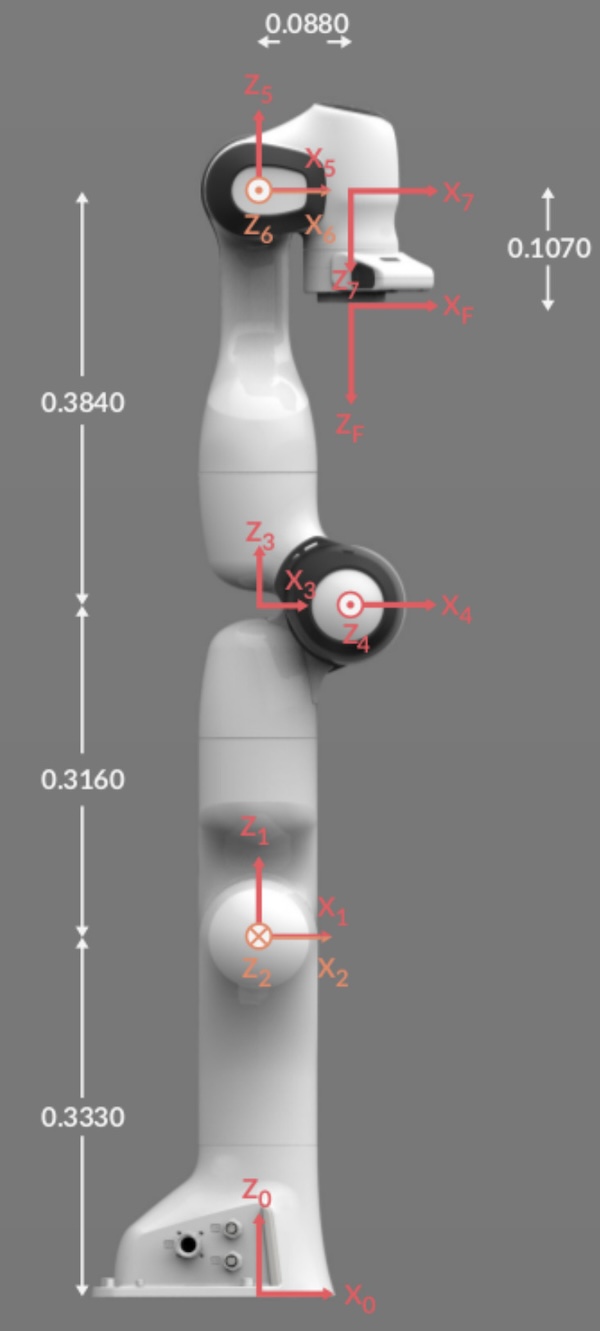}
\end{minipage}
\caption{Denavit-Hartenberg Parameters \cite{DH9646185} and image \cite{frankaemika2024} of Franka Emika robotic arm, with $q_i$ being the joint angle of the $i$th revolute joint.}
\label{fig:franks_emika_and_DH}
\end{figure}

Figure~\ref{fig:franks_emika_and_DH} shows the Franka Emika Panda robot used in this dataset and its kinematic parameters according to the Denavit-Hartenberg convention. This robot is equipped with a $7$-axis revolute joint, where the angle of these joints is defined in radian as $q_i \text{ for } i \in \{ 1, 2, ..., 7 \}$.
The Franka Emika robotic arm is a collaborative robot (cobot) designed to work safely alongside humans in various environments, ranging from industrial settings to direct interaction scenarios. Unlike conventional industrial robots, which are typically enclosed for safety reasons, the Franka Emika arm can perform tasks in close proximity to people without posing a hazard~\cite{haddadin2022franka}. This capability makes it ideal for operations that require direct physical interaction, such as drilling, screwing, polishing, and a wide range of inspection and assembly tasks.
The Franka Emika robotic arm provides a $3$ kg payload capacity and a reach of $850$ mm. The robot weighs approximately $18$ kg and its repeatability is $0.1$ mm. Repeatability is a measure of the ability of the robot to consistently reach a specified point.

\subsection*{Experiments}
\label{sec:methods:experiments}

Our dataset is composed of $60$ different primary combinations performed by the robotic arms, capturing activities through our sensor-rich modules. Our dataset encompasses four variations:

\subsubsection*{Activity}
\begin{figure}[htbp]
    \centering
    \includegraphics[width=0.6 \textwidth]{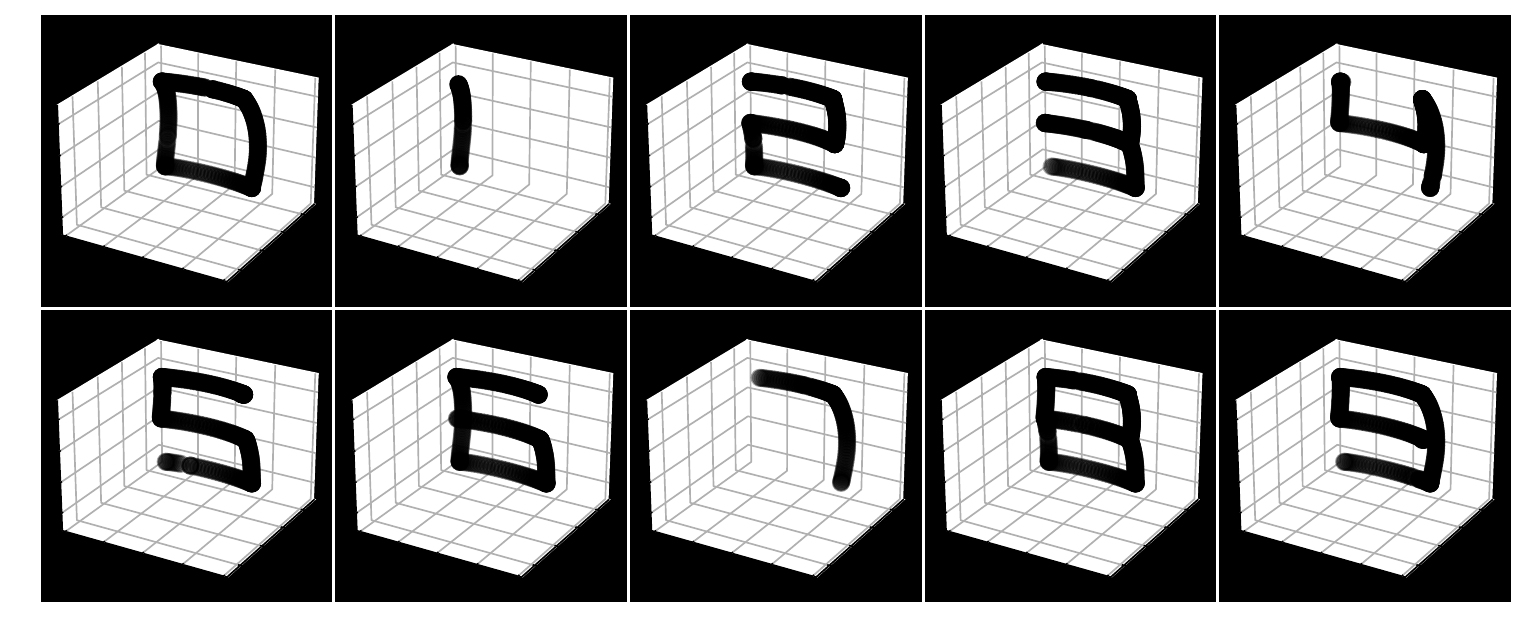}
    \caption{Numbers $0$ through $9$ are drawn by the robotic arm on a vertical imaginary plane, resulting in $10$ distinct classes of activities. The plot shows the end effector trajectories that form these numbers, with the robotic arm and background removed for clarity. For illustration purposes the initial and final parts of the robot's trajectory, where the robot positions itself from its starting point to the imaginary plane and back, are omitted.}
    \label{fig:actions}
\end{figure}

The Franka Emika robotic arms were programmed to draw the digits $0$ through $9$ on a vertical imaginary plane, creating ten distinct activity classes. The positions of the end effector for each activity are shown in Figure~\ref{fig:actions}. We denote the activities performed by the robots as $A \in \{0,1,\cdots,9\}$.

\begin{figure}[htbp]
    \centering
    \includegraphics[width=0.95 \textwidth]{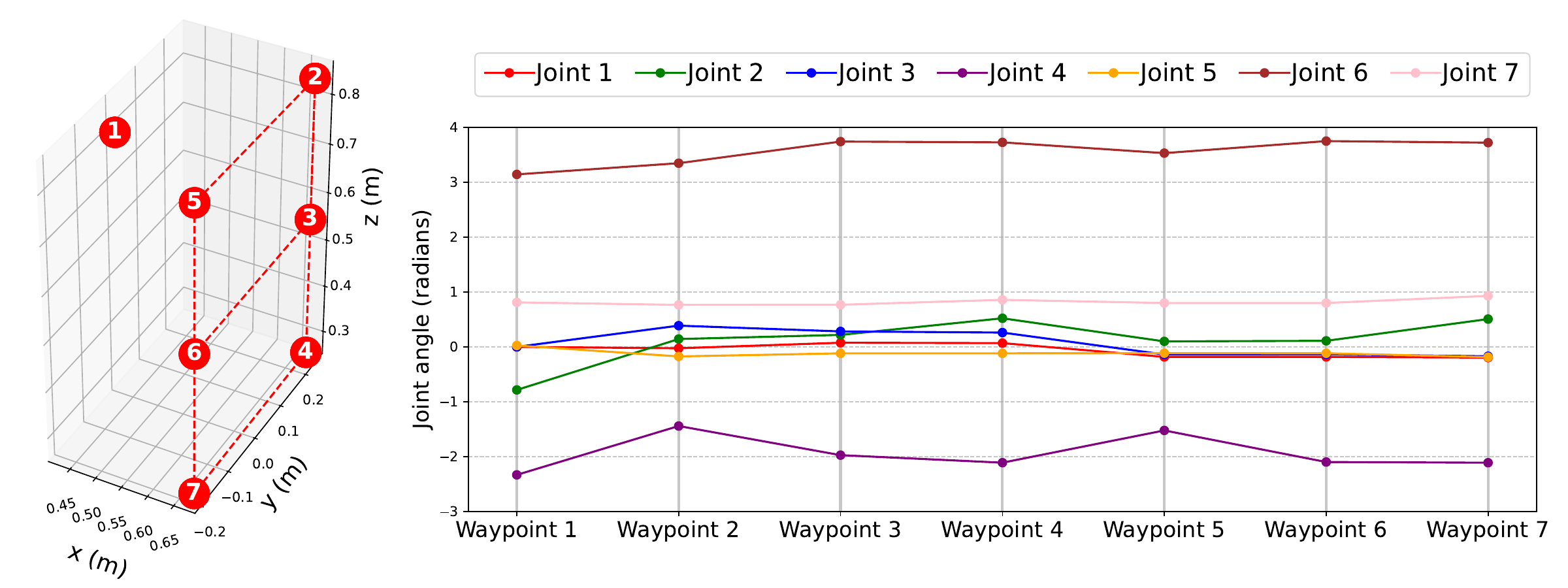}
    \caption{Each activity involves writing a digit using a robotic arm, defined by seven distinct waypoints. The left plot illustrates the waypoints in the end-effector space, while the right plot represents them in the joint space. Each digit is written by following a specific sequence of waypoints in the joint space, with the robot always starting and ending at waypoint 1. For example, to write the digit $7$, the robot follows the sequence $\{1, 2, 5, 7, 1\}$.}
    \label{fig:waypoints}
\end{figure}

To execute these activities, we generated seven fixed waypoints. These waypoints are depicted in Figure~\ref{fig:waypoints}, illustrating their positions in both the joint space and the corresponding end effector space. The activities are performed using the waypoints in the joint position space. For each activity, the robot follows an appropriate sequence of waypoints from waypoint 2 to waypoint 7. The robot always starts and ends at waypoint 1, after completing the digit in the imaginary plane.

\subsubsection*{Robot Number}
Indicated by $R \in \{1,2\}$, this specifies which of the two available robotic arms is performing the activity based on the numbering notation in Figure~\ref{fig:floor_plan}.

\subsubsection*{Robot velocity}
Denoted by $V \in \{\text{\textit{High}},\: \text{\textit{Medium}},\: \text{\textit{Low}}\}$, this describes the velocity level at which the robot performs the activity.
To enforce these velocities, we applied the following limits on the maximum joint acceleration and deceleration of the joint angles:

\begin{itemize}
    \item \textit{High}: A maximum joint acceleration and deceleration of $a_{\text{max}} = 2.5 \; \text{radian/s}^2$ have been applied on each joint in this setting. This corresponds to $50\%$ of the robot's maximum allowed acceleration.

    \item \textit{Medium}: A maximum joint acceleration and deceleration of $a_{\text{max}} = 2.0 \; \text{radian/s}^2$ have been applied on each joint in this setting. This corresponds to $40\%$ of the robot's maximum allowed acceleration.

    \item \textit{Low}: A maximum joint acceleration and deceleration of $a_{\text{max}} = 1.5 \; \text{radian/s}^2$ have been applied on each joint in this setting. This corresponds to $30\%$ of the robot's maximum allowed acceleration.
\end{itemize}

\subsubsection*{Motion uncertainty}
\label{sec:methods:motion_uncertainty}
Denoted by $U \in \mathbb{R}^+$, where $\mathbb{R}^+$ represents the positive real numbers, measures the $L_2$ norm error of the end effector's position relative to its intended ground truth trajectory over time.

\begin{figure}[htbp]
    \centering
    \includegraphics[width=0.95 \textwidth]{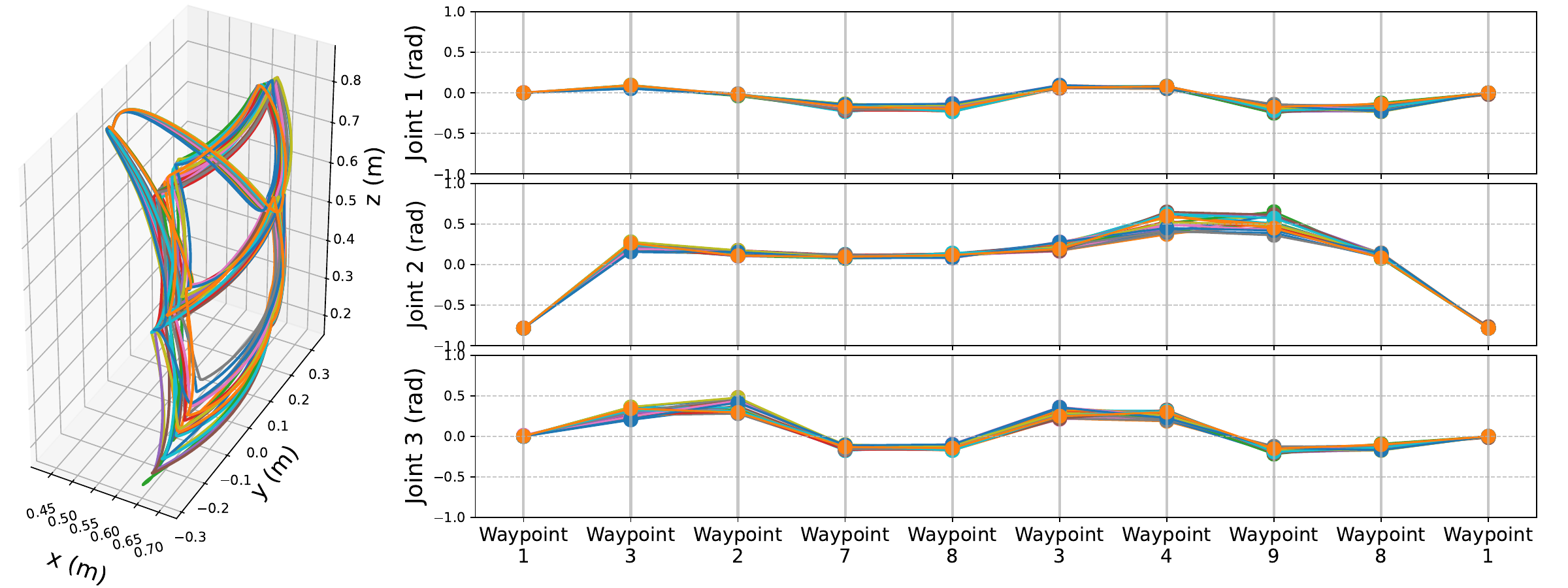}
    \caption{The effect of uncertainty on the end effector position (left) and on the joint positions (right) of the robot. Each color represents the result of a repetition where the robot performs the activity with $A=8$, $R=1$, and $V=\text{\textit{High}}$. Only the first three joints are shown in this figure, as no uncertainty is applied to the remaining joints. Additionally, waypoint 1 has no added uncertainty, ensuring that the robot always starts and ends in the same position.}
    \label{fig:uncertainty_explain}
\end{figure}

\begin{figure}[htbp]
    \centering
    \includegraphics[width=0.95 \textwidth]{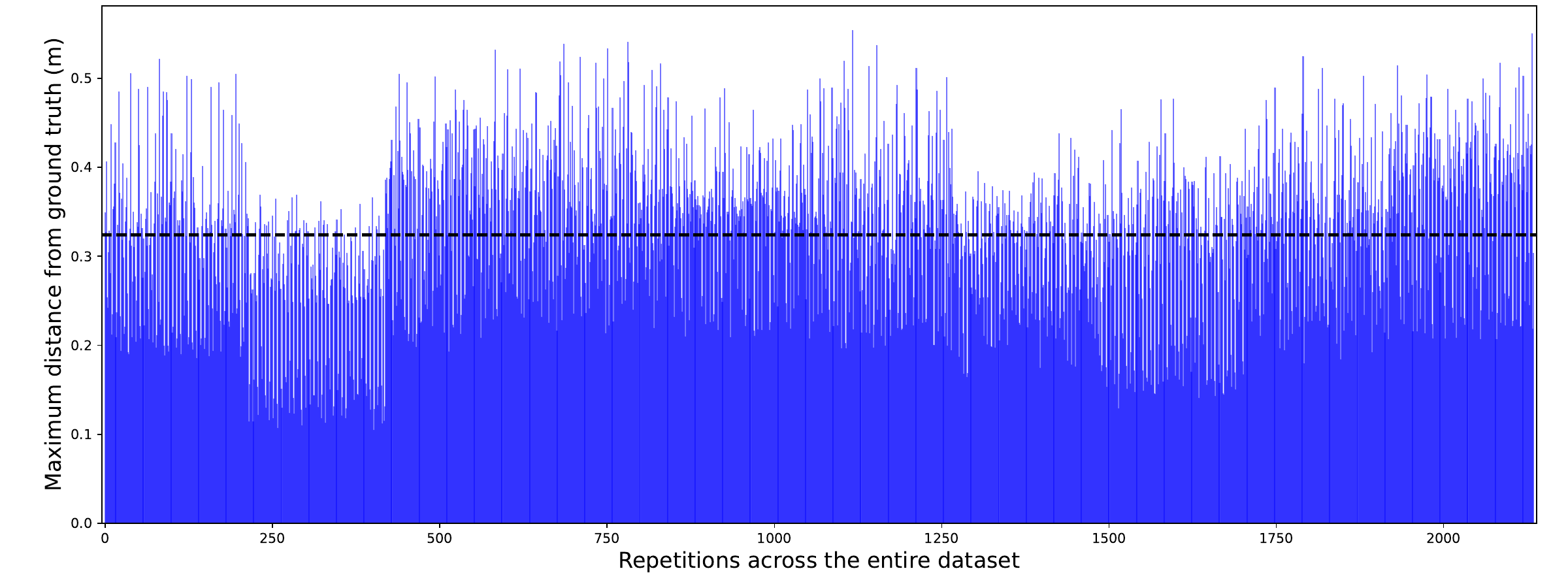}
    \caption{The maximum distance of the robots' end effector from the ground truth across all repetitions in the dataset. The dashed black line indicates the average distance.}
    \label{fig:repeatability_end_effector}
\end{figure}

The Franka Emika robotic arm repeatability is $0.1$ mm, which ensures that the robot performs a very similar motion for each activity, to make the dataset realistic We manually and purposefully introduced a layer of uncertainty into the robots' motion to ensure that different repetitions in our dataset are not identical. This approach simulates the variability of handwritten digits performed by humans, which are inherently non-identical. To achieve this, we added multiplicative uniform noise $\epsilon \sim \mathcal{U}(0.7, 1.3)$ to the first three joints of the robot ($q_1, q_2, q_3$) for each waypoint in the sequence, except for the first waypoint. This ensures that the start and end positions of the robot remain consistent. We chose the first three joints because they have the most significant effect on the robots' final trajectory, while the remaining four joints primarily influence the orientation of the robot's end effector. Figure~\ref{fig:uncertainty_explain} illustrates the impact of these uncertainties on the first three joints and the end effector position of the robot in different repetitions on the same activity. Using this approach, we have increased the robot's repeatability to an average of $32$ cm. To illustrate this, Figure~\ref{fig:repeatability_end_effector} presents the maximum distance between the robot's trajectory and the ground truth across all repetitions in the dataset.

\begin{figure}[htbp]
    \centering
    \includegraphics[width=0.92 \textwidth]{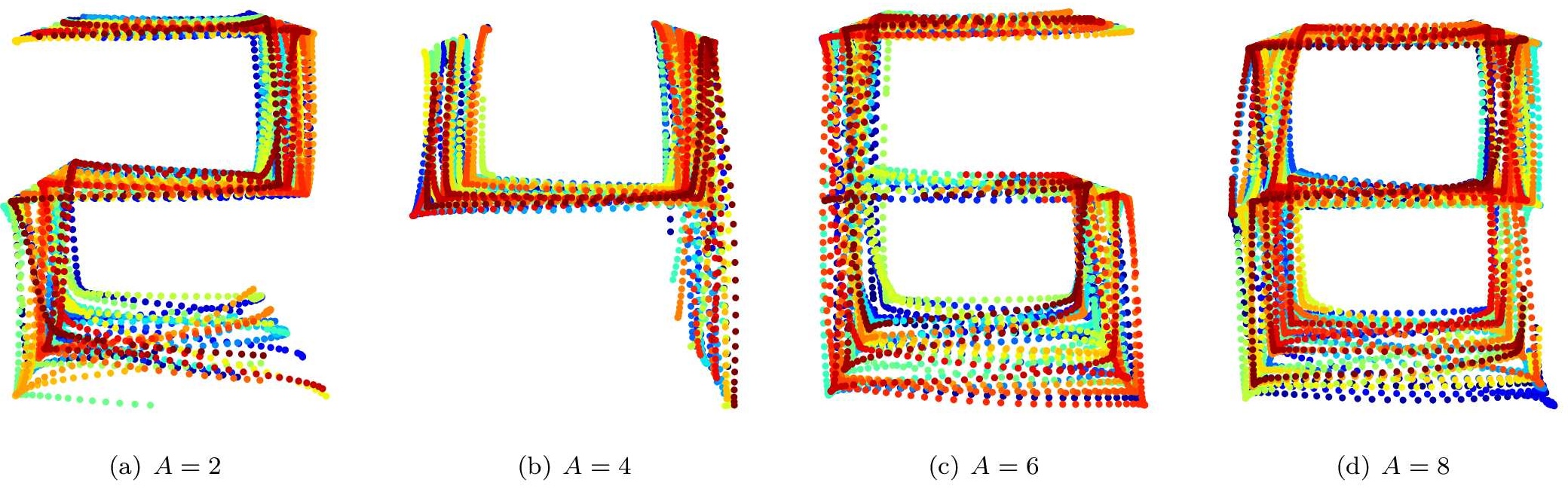}
    \caption{Motion uncertainty of all the repetitions in four of the primary combinations as a sample. All with $R = 1$, and $V = \text{\textit{High}}$, but with different values of $A$. For illustration purposes, the path is projected on a $2$D imaginary plane and the initial and final parts of the robot's trajectory, where the robot positions itself from its starting point to the imaginary plane and back, are omitted.}
    \label{fig:colorful_uncertainty}
\end{figure}

The combination of ten activities, two robots performing these activities, and three velocity levels results in a total of $60$ unique primary combinations. For each primary combination, we have collected $32$ repetitions. Each repetition spans $15$ seconds, during which the robot performs the action with consistent variations in activity, robot arm, and velocity, while incorporating motion uncertainty. This introduces deviations in each repetition as the robot writes on an imaginary plane, adding a realistic layer of complexity to the dataset. Figure~\ref{fig:colorful_uncertainty} shows the uncertainty in motion of all the repetitions in four of the primary combinations as a sample projected on a $2$D imaginary plane.

\subsection*{WiFi CSI Modality}
\label{sec:methods:csi_modality}
As wireless signals propagate, they encounter various obstacles in the environment, leading to reflections and scattering, a phenomenon known as multipath fading~\cite{yang2013rssi}. WiFi CSI facilitates the analysis of subcarrier propagation from the transmitter to the receiver in wireless communications~\cite{wang2019survey}. The channel model is represented as

\begin{equation}
\mathbf{y} = \mathbf{Hx} + \boldsymbol{\eta},
\end{equation}
where $\mathbf{x}$, $\mathbf{y}$, and $\boldsymbol{\eta}$ denote the transmitted signal vector, received signal vector, and additive noise vector, respectively~\cite{wang2019survey}. The channel matrix $\mathbf{H} \in \mathbb{C}^{T\times S}$ encapsulates the characteristics of the wireless channel, including multipath propagation, fading, and other impairments, and is defined as

\begin{equation}
    \mathbf{H}=\left[\begin{array}{cccc}h_1[1] & h_2[1] & \ldots & h_S[1] \\ h_1[2] & h_2[2] & \ldots & h_S[2] \\ \vdots & \vdots & \ddots & \vdots \\ h_{1}[T] & h_{2}[T] & \ldots & h_{S}[T]\end{array}\right],
    \label{eq:CSI_matrix}
\end{equation}
where $S$ and $T$ represent the number of subcarriers for each antenna and the number of transmitted packets, respectively. Each element of the matrix $\mathbf{H}$ corresponds to a complex value, known as the channel frequency response, and is given by

\begin{equation}
h_s[t] = a_s e^{j\phi_s},
\end{equation}
where $a_s$ and $\phi_s$ denote the amplitude and phase of subcarrier $s$ at timestamp $t$, respectively. For human activity recognition (HAR)\cite{salehinejad2022litehar,yousefi2017surv,widar2019} and robot activity recognition (RAR)\cite{zandi2023robot, zandi2024robofisense, zandi2024enhancing}, studies primarily focus on $\mathbf{A}\in \mathbb{R}^{T\times S}$, which corresponds to the element-wise amplitude of $\mathbf{H}$, disregarding the phase component.

\begin{figure}[htbp]
    \centering
    \includegraphics[width=0.95 \textwidth]{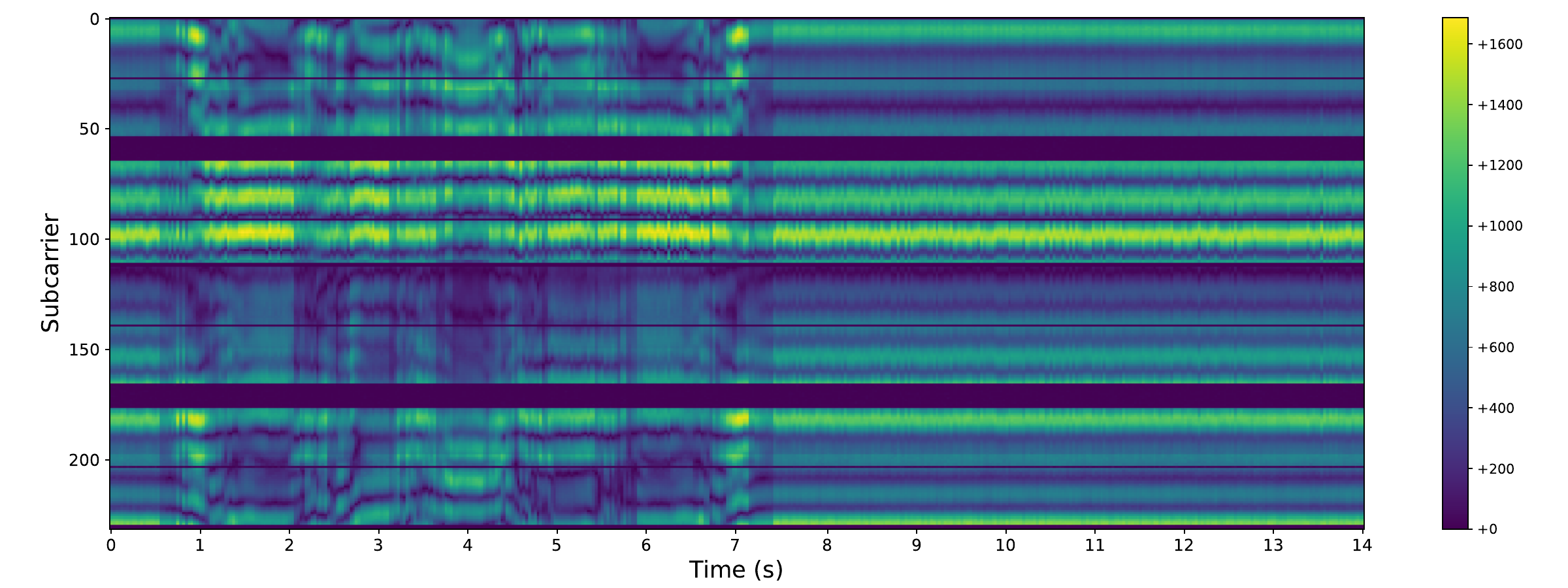}
    \caption{Plot of CSI amplitudes, in a repetition with $A = 0$, $R = 1$, and $V = \text{\textit{High}}$, while $M = 1$ module capturing the CSI measurements.}
    \label{fig:csi_plot}
\end{figure}

For each $15$-second repetition, we collected CSI measurements at a $30$ Hz frequency, over an $80$ MHz bandwidth which gives $256$ number of subcarriers at each time stamp. This resulted in a $450 \times 256$ complex matrix for each sensor-rich module. These matrices are stored in a \textit{json} file. Additionally, we included received signal strength (RSS) information for each timestamp as well. Figure~\ref{fig:csi_plot} displays a sample plot of the amplitudes of a CSI matrix.

\subsection*{Video Modality}
\label{sec:methods:vision_modality}
For each $15$-second repetition, we collected video measurements at a frequency of $30$ Hz, synchronized with the CSI measurements, using three sensor-rich modules, each containing a stereo camera. For each sample, three ZED $2$ stereo cameras simultaneously recorded RGB videos at a resolution of $2560 \times 720$ pixels, with the frames from the left and right lenses of each stereo camera horizontally concatenated. This setup allowed us to capture three different views of the same action. Each camera, equipped with two lenses, provided stereo video, resulting in a dataset that includes $3 \text{ (cameras)} \times 2 \text{ (lenses)} = 6$ videos for each repetition.

\subsection*{Audio Modality}
\label{sec:methods:audio_modality}
Audio signals are continuous waveforms that represent sound waves in a format that can be processed by digital systems. These waveforms are characterized by frequency, amplitude, and phase, which contain rich information about the environment and the sources of sound. In the context of activity recognition, audio signals offer a non-intrusive and cost-effective means to infer activities and interactions. By analyzing the acoustic patterns and variations over time, it is possible to identify specific activities based on the distinct sounds associated with each activity. Advanced machine learning algorithms, particularly those leveraging deep learning, have demonstrated significant success in classifying and recognizing activities from audio data. These algorithms extract useful information, such as the frequency and amplitude of the sound wave over time, to analyze and predict activities \cite{reinolds2022deep, stork2012audio}. 

In this paper, we employ auditory perception as many robot activities produce characteristic sounds from which we can effectively infer corresponding actions. We view audio not as a replacement but as a complement to existing sensory modalities. By fusing audio with other sensory data, we aim to achieve particularly robust activity recognition across a wide range of conditions. This multimodal approach enhances the accuracy and reliability of activity recognition systems, making them more effective in diverse environments.

\begin{figure}[htbp]
    \centering
    \includegraphics[width=0.95 \textwidth]{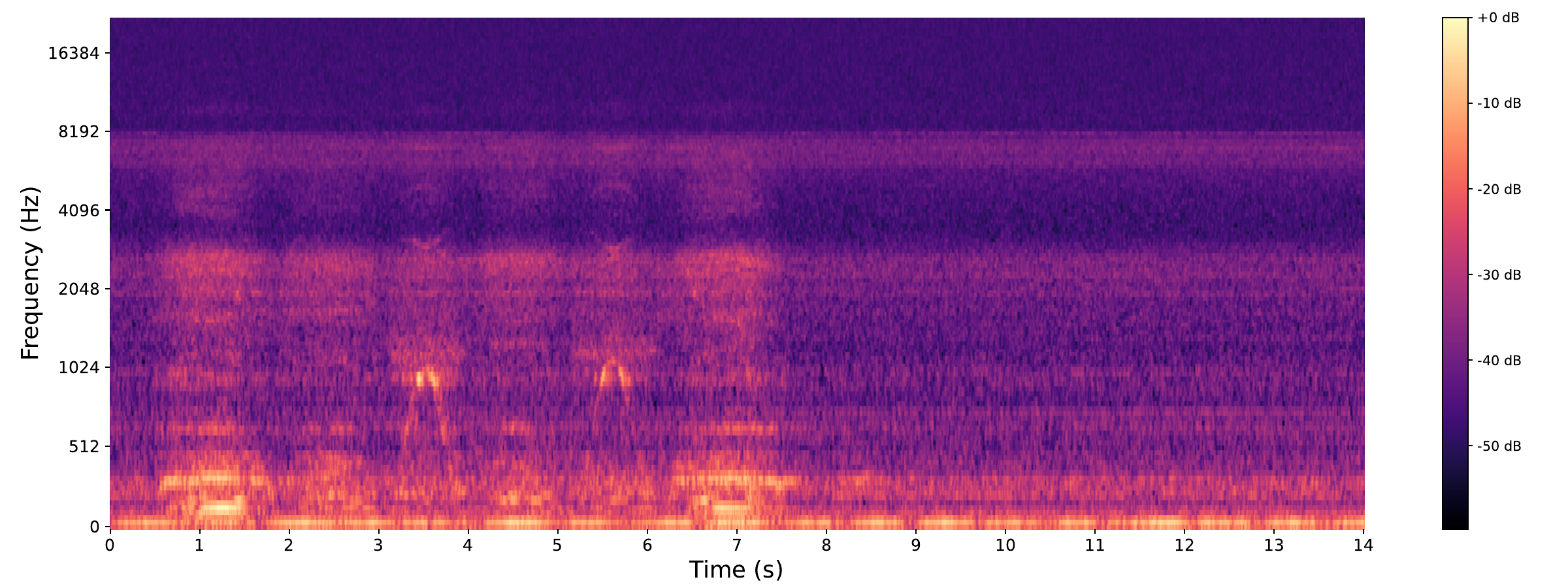}
    \caption{Plot of audio spectrogram, in a repetition with $A = 0$, $R = 1$, and $V = \text{\textit{High}}$, with $M = 1$ module capturing the audio.}
    \label{fig:audio_spectrogram_plot}
\end{figure}

For each $15$-second repetition, we collected audio measurements from each sensor-rich module at a sampling rate of $44{,}100$~Hz while the start and end times were synchronized with our other modalities.
 To analyze the audio data, we preprocess the audio files by computing their spectrograms. While we delve more into the calculation of the audio spectrogram in the Technical Validation section, a sample spectrogram plot of one of the captured audio is shown in Figure~\ref{fig:audio_spectrogram_plot}.

\subsection*{True Trajectory}
\label{sec:methods:true_trajectory}
\begin{figure}[htbp]
    \centering
    \includegraphics[width=0.5 \textwidth]{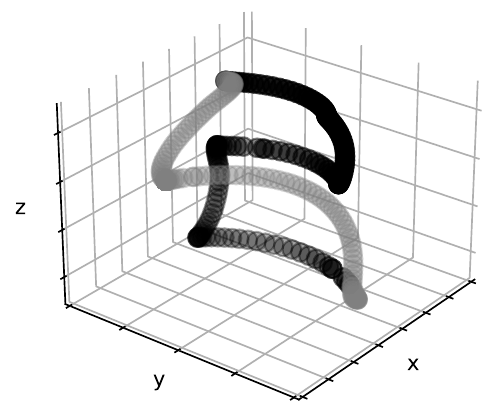}
    \caption{Plot of the end effector's position of the robot during a repetition with $A = 2$, $R = 1$, and $V = \text{\textit{High}}$. The black trajectory represents the robot drawing the number on an imaginary plane, while the gray trajectory illustrates the robot's movement from a fixed starting position to the imaginary plane and back.}
    \label{fig:end_effector_position_plot}
\end{figure}

For each $15$-second repetition, we provided the true joint positions of the robots at a frequency of $30$ Hz, synchronized with the CSI and video measurements. At each timestamp, each joint position, consisting of $q_1$ to $q_7$, corresponding to the $7$-axis revolute joints of the robot, has been stored in radians in a \textit{json} file. For convenience, the $x-y-z$ position of the end effector (EE) in the Cartesian coordinate system is also provided, presented in the robot frame aligned with the $x_0$ and $z_0$ axes and following the right-hand rule, as shown in Figure~\ref{fig:franks_emika_and_DH}. Figure~\ref{fig:end_effector_position_plot} displays a sample plot of the end effector's Cartesian position of one of the robots performing an activity.

\subsection*{Synchronization}
\label{sec:methods:synchronization}
During data collection, each module incorporates local timestamping directly on the hardware. The timestamped data is subsequently transmitted across the network for further processing. Although this configuration effectively captures data from individual modules, synchronizing timestamps is critical for deployments involving multiple modules. To address this challenge, we developed a method whereby packets containing CSI, video, and audio data, each with its own timestamp, are redirected to a specialized system known as the monitor.

\begin{figure}[htbp]
    \centering
    \includegraphics[width=0.6 \textwidth]{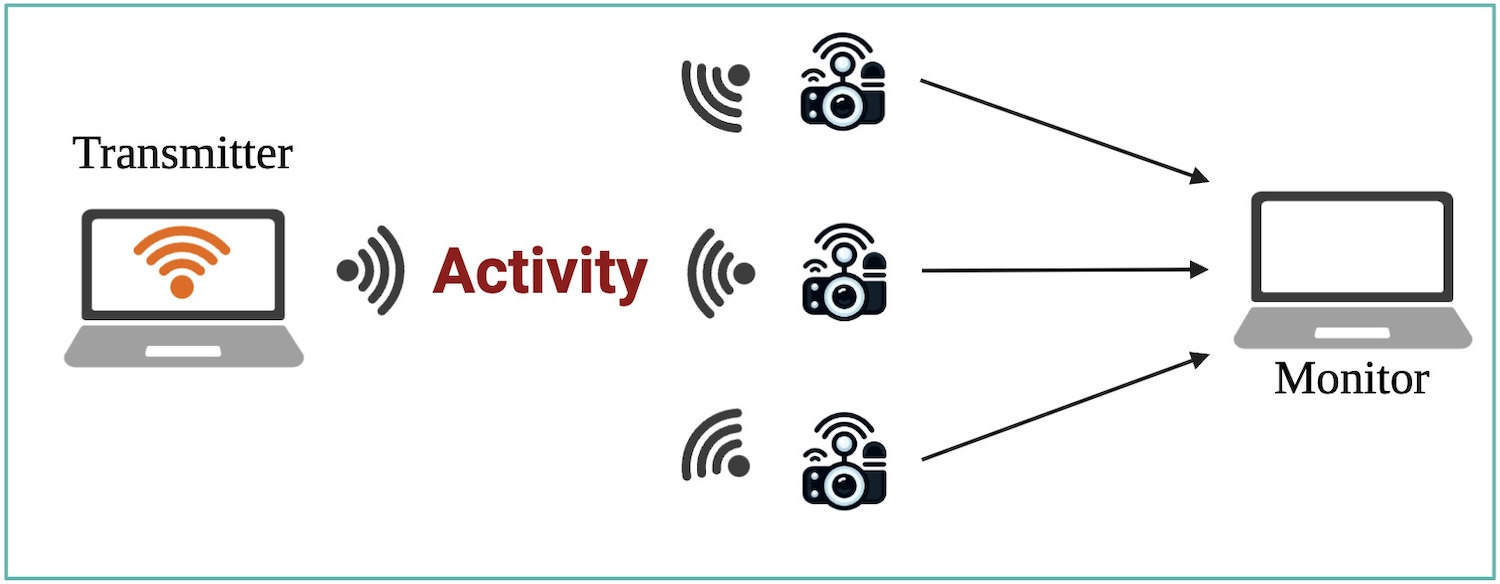}
    \caption{Communication setup between modules and the monitor. The modules will gather the data and transfer them using a wire connection to the monitor for further processing, timestamping, and synchronization during the data collection procedure.}
    \label{fig:communication_setup}
\end{figure}

The monitor serves as a central hub, collecting packets from various modules and assigning synchronized timestamps to the data. A visual depiction of this intercommunication process is presented in Figure~\ref{fig:communication_setup}. Notably, according to the Nexmon project's specifications, Raspberry Pis configured as CSI sniffers forfeit their WiFi communication capabilities. To overcome this restriction, we connected the sniffers and the monitor using Ethernet cables, thus ensuring uninterrupted communication between them.

\begin{figure}[htbp]
    \centering
    \includegraphics[width=0.8 \textwidth]{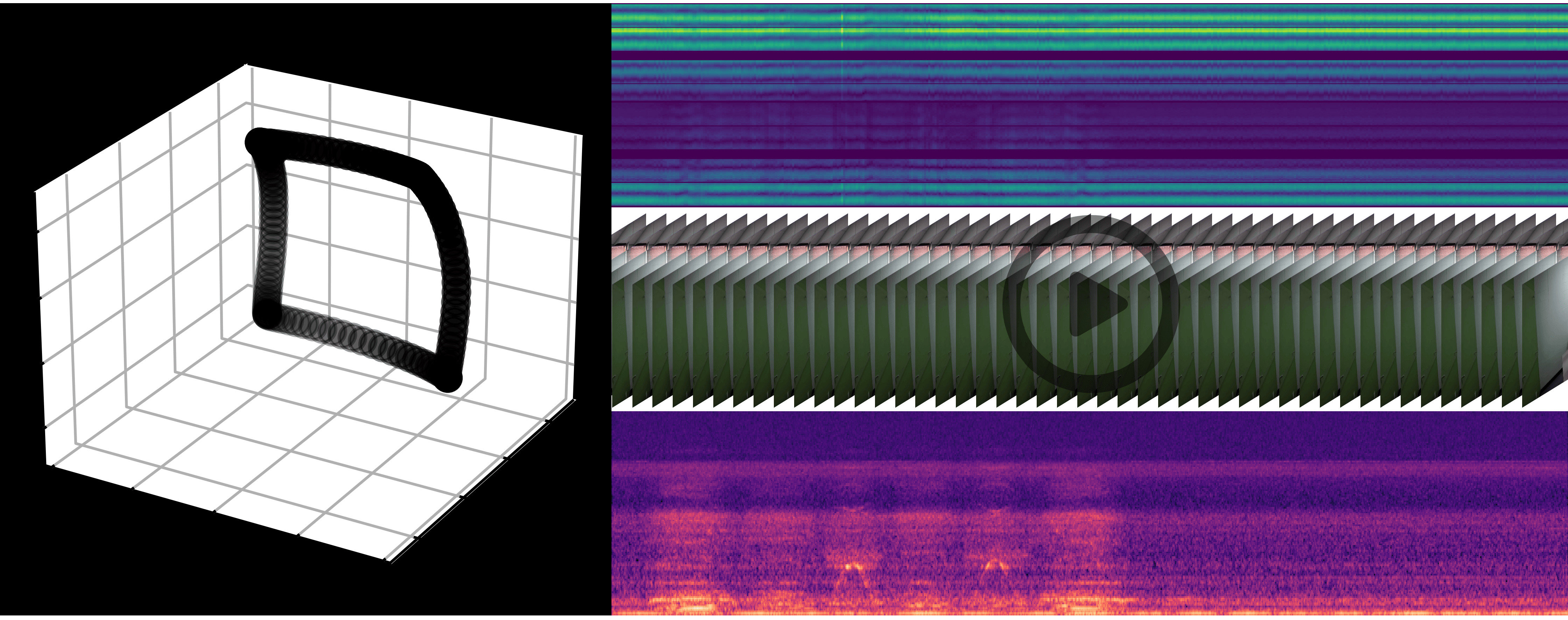}
    \caption{Plot of the three modalities and the robot's true trajectory in a repetition with $A = 0$, $R = 1$, and $V = \text{\textit{High}}$, captured by module $M = 1$. The video, CSI, true trajectory, and start and end time of the audio are synchronized. For illustration purposes, the initial and final parts of the robot's movement, where it positions itself from its starting point to the imaginary plane and back, are omitted.}
    \label{fig:all_in_one}
\end{figure}

Using this configuration, the CSI, video, the true trajectories of the robots, and the start and end time of the audio are synchronized with each other and between all the sensor-rich modules, providing a comprehensive dataset for multi-modal passive MRAR. Figure~\ref{fig:all_in_one} compactly shows different synchronized modalities in an experiment.

\begin{figure}[htbp]
    \centering
    \includegraphics[width=0.96 \textwidth]{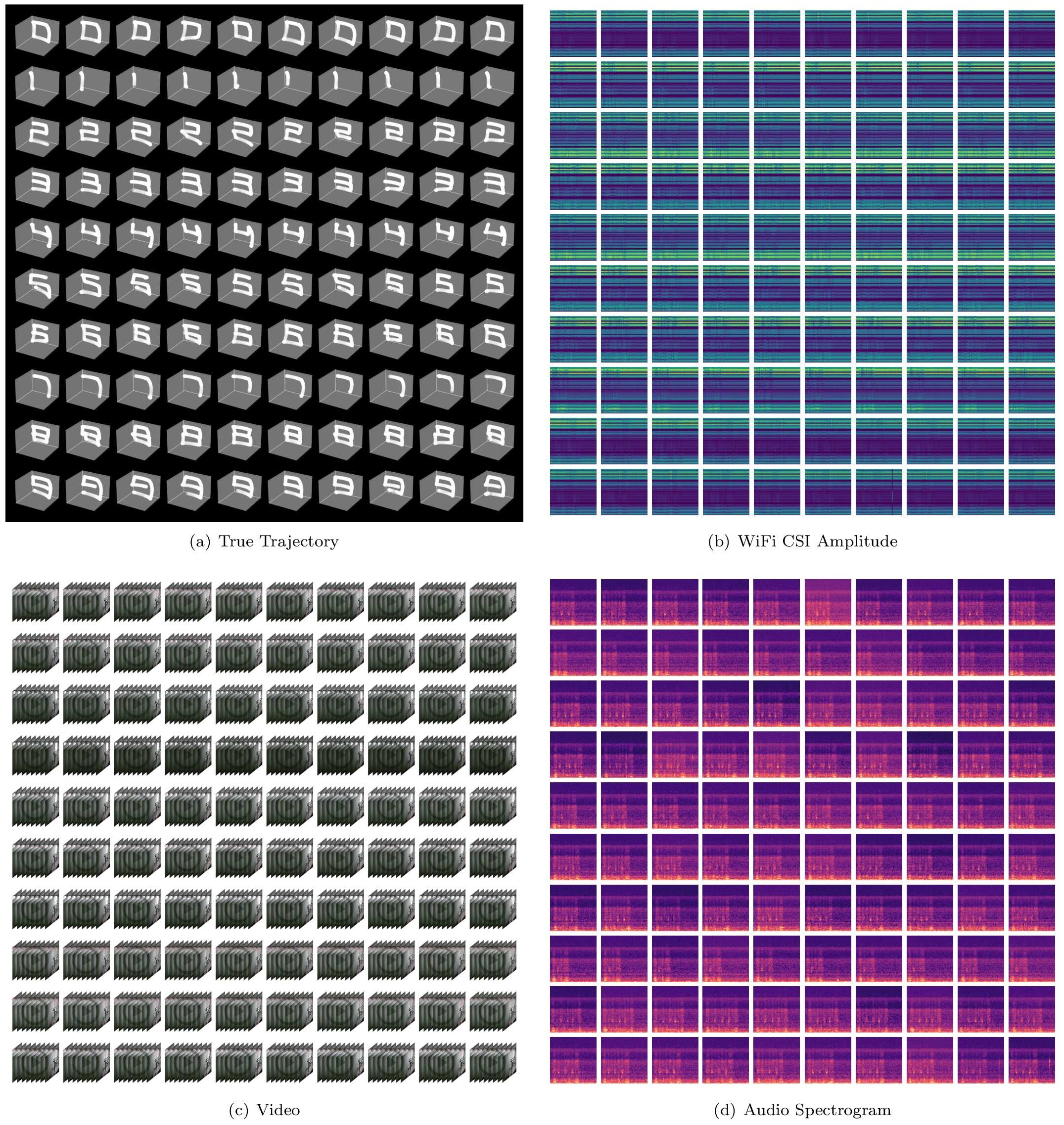}
    \caption{MNIST formatted plots of the true trajectory, WiFi CSI amplitude, video, and audio spectrogram. Each row represents activities from $0$ to $9$ and the columns are different repetitions in the same primary combination.}
    \label{fig:plot_like_mnist}
\end{figure}

Figure~\ref{fig:plot_like_mnist} shows MNIST formatted plots of the $10$ activities and different repetitions based on our modalities.

\section*{Data Records}
\label{sec:data_records}
The dataset is available for download from our Figshare repository \cite{figshare_dataset_2024}.

\begin{figure}[htbp]
    \centering
    \includegraphics[width=0.90 \textwidth]{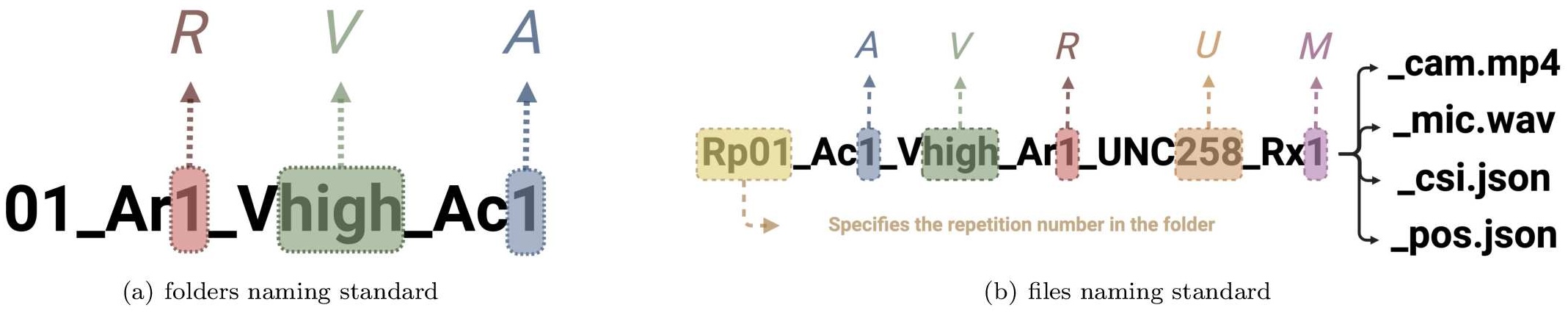}
    \caption{Naming standard for folders and files in our dataset. The naming standard for folders, dedicated to each primary combination, specifies the robot performing the activity (denoted by $R$), the velocity of the activity (denoted by $V$), and the specific activity being performed (denoted by $A$). The naming standard for each file within a folder shares the same values for $A$, $V$, and $R$, and varies by the robot's motion uncertainty (denoted by $U$). The value of $U$ is always a floating-point number with two decimal places. For example, UNC258 represents $U = 2.58$. The Rx indicates the module from which the data is sourced; it can be a specific number or "all," where "all" signifies that data from all sensors/robots are collected in the same file.}
    \label{fig:naming_standard}
\end{figure}

Based on variations in activity, robot, and velocity, we have $60$ primary combinations. Each combination has a dedicated folder in the dataset, named according to the standard described in Figure~\ref{fig:naming_standard}(a). Within each primary combination, there are at least $32$ repetitions, all with $15$-second duration and specific $R$, $V$, and $A$ variations, differing only by assigned motion uncertainty. Each repetition contains $8$ files as listed in Table~\ref{table:repetition_files}, following the naming convention outlined in Figure~\ref{fig:naming_standard}(b).

\begin{table}[htbp]
\centering
\begin{tabular}{|l|l|}
\hline
Data & Associated files \\
\hline
CSI & One \textit{json} file containing CSI data for all modules\\
\hline
Video & Three \textit{mp4} files containing video data for each module\\
\hline
Audio & Three \textit{wav} files containing audio data for each module\\
\hline
True robot trajectory & One \textit{json} file containing trajectory data for both robots\\
\hline
\end{tabular}
\caption{Associated files and their formats corresponding to each repetition of a primary combination.}
\label{table:repetition_files}
\end{table}

\subsection*{WiFi CSI Description}
This section describes the structure of the data files residing in the CSI files, which are the files ending with \textit{csi.json}. Each repetition is represented by a single CSI file, which contains the CSI data for all the three sensor-rich modules. These CSI files are in \textit{json} format and consist of an array of three \textit{json} objects. Each \textit{json} object corresponds to one of the sensor-rich modules. Within each \textit{json} object, the data is organized as a set of key-value pairs, as detailed in Table~\ref{table:csi_json_description}.

\begin{table}[htbp]
\centering
\begin{tabularx}{\textwidth}{|p{2.5cm}|p{5cm}|X|}
\hline
Key & Type & Value \\
\hline
module\_number & Integer, either $1$, $2$, or $3$ & The module number based on Figure~\ref{fig:floor_plan}\\
\hline
time\_stamp\_ns & Array of integer numbers & The timestamps of CSI collections in nanoseconds\\
\hline
complex\_csi & Matrix of complex numbers & The CSI matrix as defined in eq.~\ref{eq:CSI_matrix}, where each row corresponds to a timestamp\\
\hline
RSS & Array of integer numbers & The RSS values at each timestamp\\
\hline
\end{tabularx}
\caption{Description of key-value sets of \textit{json} objects in CSI files.}
\label{table:csi_json_description}
\end{table}

\subsection*{Video Description}
This section describes the structure of the data files residing in the video files, identified by the \textit{cam.mp4} extension. Each repetition corresponds to $3$ video files, each for one of the sensor-rich modules. Each video file is in \textit{mp4} format, where each frame consists of the horizontally concatenated left and right frames of the lenses of the stereo camera, resulting in a final frame with dimensions of $2560 \times 720$. Each frame is synchronized with the timestamps provided in the CSI file of the same repetition.

\subsection*{Audio Description}
This section describes the structure of the data files residing in the audio files, identified by the \textit{mic.wav} extension. Each repetition corresponds to three audio files, each file for one of the sensor-rich modules. Each audio file is in \textit{wav} format and has been recorded for the $15$-second duration of the repetition.

\subsection*{True Robot Trajectory Description}
This section describes the structure of the data files in the position files, which have the \textit{pos.json} extension. Each repetition corresponds to one position file containing the position information of both robots. Even though only one robot is moving in each repetition, as indicated by the file name, we have included the position data for both robots. Each position file is in \textit{json} format and consists of an array of two \textit{json} objects, one for each robot. Each \textit{json} object is a set of key-value pairs, as detailed in Table~\ref{table:position_json_description}.

\begin{table}[htbp]
\centering
\begin{tabularx}{\textwidth}{|p{2.5cm}|p{5cm}|X|}
\hline
Key & Type & Value \\
\hline
robot\_number & Integer, either $1$ or $2$ & The robot number based on Figure~\ref{fig:floor_plan}\\
\hline
time\_stamp\_ns & Array of integer numbers & The timestamps of robot positions in nanoseconds (synchronized with CSI timestamps)\\
\hline
joint\_positions & Matrix of floating point numbers & The joint positions of the robot in radian, where each row corresponds to a timestamp\\
\hline
EE\_positions & Matrix of floating point numbers & The Cartesian positions of the robot's end effector in meter, where each row corresponds to a timestamp\\
\hline
\end{tabularx}
\caption{Description of key-value sets of \textit{json} objects in position files.}
\label{table:position_json_description}
\end{table}

\section*{Technical Validation}
In this section, we validate our dataset across various aspects, including robot configuration, uncertainty, and the synchronization of CSI, video and audio.

\subsection*{Joint Positions}
The documentation of the Franka Emika robot~\cite{franka_fci_specs} provides the safety specifications for the joint positions of each of the seven joints. The safe operating range for the joints is summarized in Table~\ref{table:joint_position_validation}.

\begin{table}[htbp]
\centering
\begin{tabular}{|c|c|c|c|c|c|c|c|}
\hline
\textbf{Name} & \textbf{Joint 1} & \textbf{Joint 2} & \textbf{Joint 3} & \textbf{Joint 4} & \textbf{Joint 5} & \textbf{Joint 6} & \textbf{Joint 7} \\ \hline
$q_{\text{max}}$ & 2.8973 & 1.7628 & 2.8973 & -0.0698 & 2.8973 & 3.7525 & 2.8973 \\ \hline
$q_{\text{min}}$ & -2.8973 & -1.7628 & -2.8973 & -3.0718 & -2.8973 & -0.0175 & -2.8973 \\ \hline
\end{tabular}
\caption{Safe operating ranges for the joint positions of the Franka Emika robot in radians.}
\label{table:joint_position_validation}
\end{table}

\begin{figure}[htbp]
    \centering
    \includegraphics[width=0.95 \textwidth]{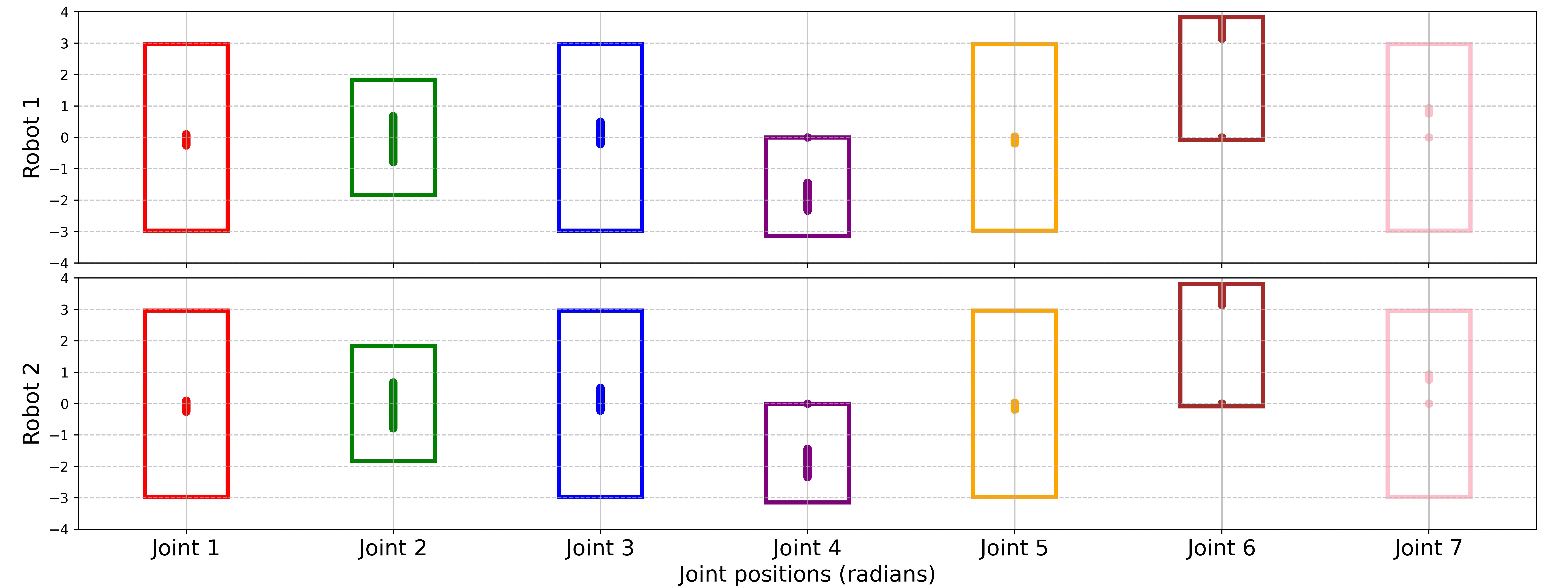}
    \caption{The joint positions for each robot are shown using filled circles, with each joint represented separately in radians. For each joint, all positions used throughout the dataset are displayed. The boxes indicate the maximum and minimum values allowed based on the specifications of the robots. Given the similarity in actions performed by both robots, their joint position ranges are nearly identical and closely aligned.}
    \label{fig:joint_position_validation}
\end{figure}

Figure~\ref{fig:joint_position_validation} shows the joint positions, in radians, used throughout our entire dataset for both robots separately. As evident from the figure, all joint positions in the dataset fall within the safe operating range of the robots.

\subsection*{Joint Velocities}
The documentation of the Franka Emika robot also provides the safety specifications for the joint velocities of each of the seven joints. The safe operating velocity ranges for the joints are summarized in Table~\ref{table:joint_velocity_validation}.

\begin{table}[htbp]
\centering
\begin{tabular}{|c|c|c|c|c|c|c|c|}
\hline
\textbf{Name} & \textbf{Joint 1} & \textbf{Joint 2} & \textbf{Joint 3} & \textbf{Joint 4} & \textbf{Joint 5} & \textbf{Joint 6} & \textbf{Joint 7} \\ \hline
$\dot{q}_{\text{max}}$ & 2.1750 & 2.1750 & 2.1750 & 2.1750 & 2.6100 & 2.6100 & 2.6100 \\ \hline
\end{tabular}
\caption{Safe operating velocity ranges for the joints of the Franka Emika robot.}
\label{table:joint_velocity_validation}
\end{table}

\begin{figure}[htbp]
    \centering
    \includegraphics[width=0.95 \textwidth]{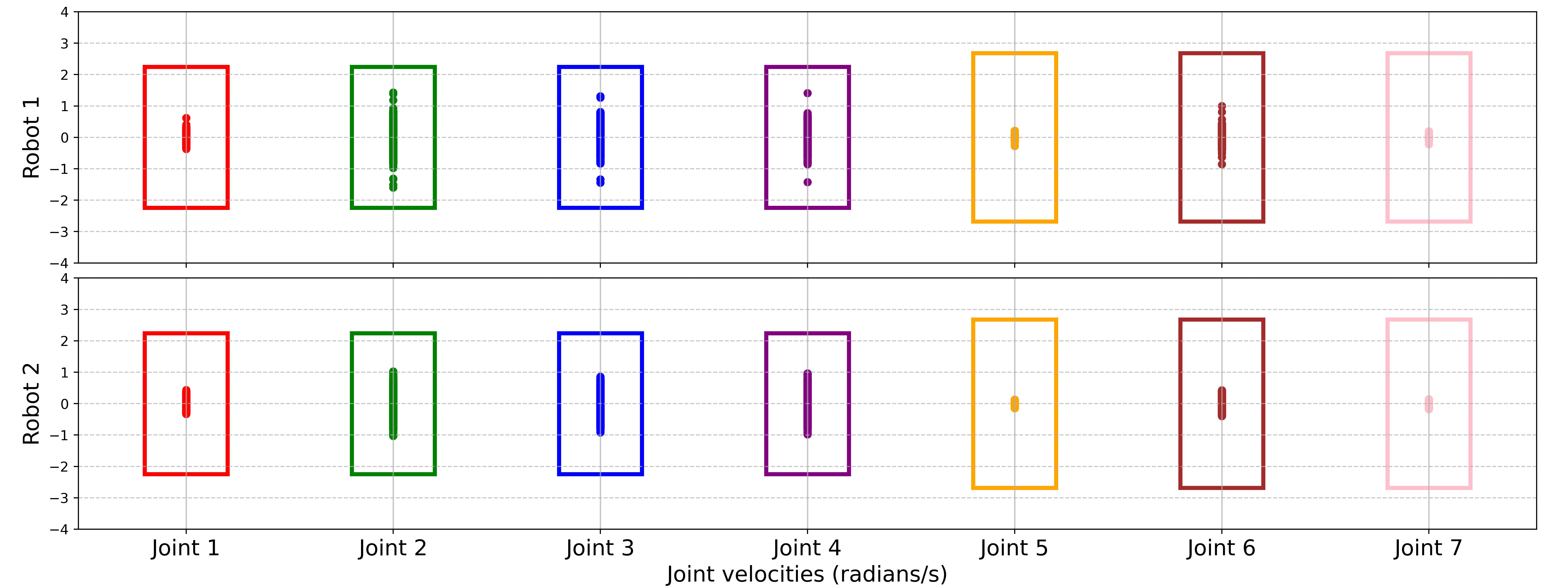}
    \caption{The joint velocities for each robot are shown using filled circles, with each joint represented separately in radian/s. For each joint, all velocities used throughout the dataset are displayed. The boxes represent the maximum and minimum values allowed according to the specifications of the robots.}
    \label{fig:joint_velocity_validation}
\end{figure}

Figure~\ref{fig:joint_velocity_validation} shows the joint velocities, in radian/s, used throughout our entire dataset for both robots separately. As evident from the figure, the joint velocities in the dataset fall within the safe operating range of the robots.

\begin{figure}[htbp]
    \centering
    \includegraphics[width=0.95 \textwidth]{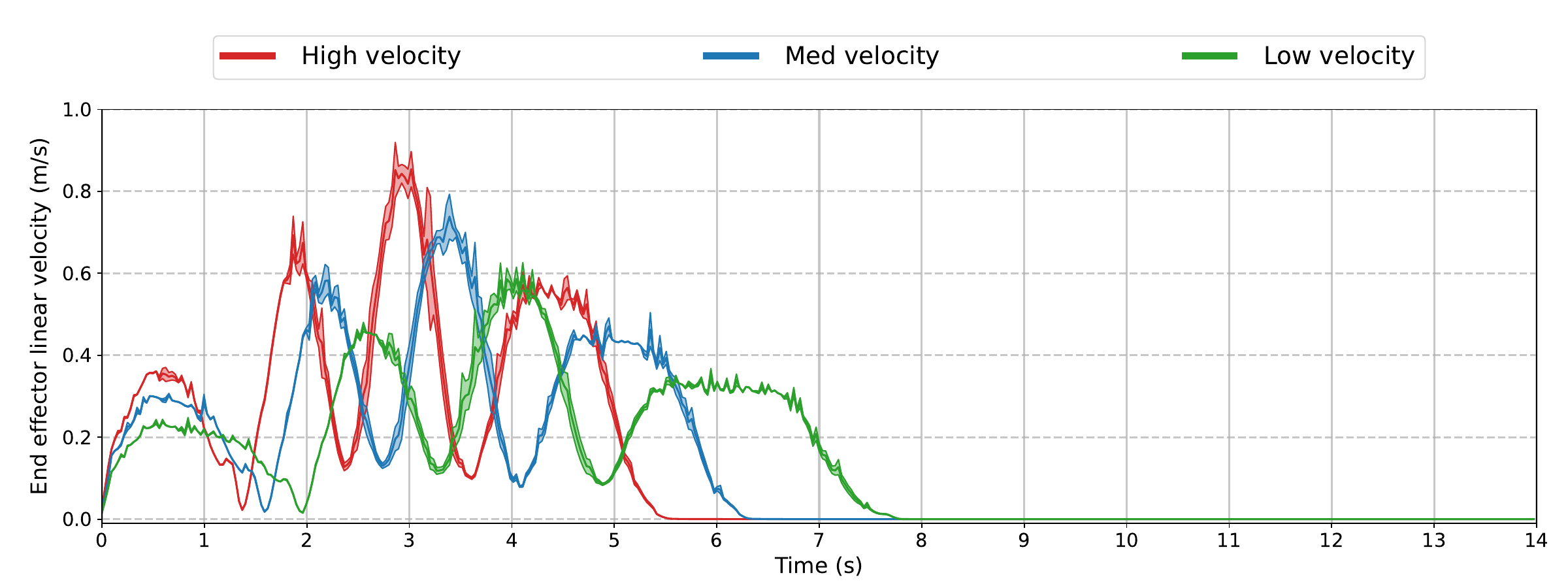}
    \caption{The linear velocity of the end effector over time for different velocity levels: \textit{High}, \textit{Medium}, and \textit{Low}. This comparison is based on repetitions with $A=7$ and $R=1$. The solid lines represent the average velocity across all repetitions, while the shaded areas around them indicate the variance. As expected, the \textit{High} velocity is consistently greater than the \textit{Medium} velocity, and the \textit{Medium} velocity exceeds the \textit{Low} velocity. In $A=7$, the robot follows five waypoints, corresponding to four motions between these waypoints. Each peak in the plot represents one of these motions. It is clear that the different velocity levels behave as expected, with the \textit{High} velocity showing higher peaks compared to \textit{Medium}, and \textit{Medium} showing higher peaks compared to \textit{Low}. Additionally, the \textit{High} velocity completes the activity faster than \textit{Medium}, and \textit{Medium} finishes sooner than \textit{Low}.}
    \label{fig:joint_velocity_compare}
\end{figure}

Figure~\ref{fig:joint_velocity_compare} shows the linear velocity of the robot's end effector for the different velocity levels used in this dataset. It illustrates how the \textit{High}, \textit{Medium}, and \textit{Low} velocity levels affect the motion of the end effector. As expected, the velocity is higher when using the \textit{High} setting compared to the \textit{Medium} setting, resulting in the activity being completed more quickly. A similar pattern is observed when comparing the \textit{Medium} and \textit{Low} velocity levels. 
When comparing the velocity levels, it is important to note that, for instance, the \textit{Medium} velocity exhibits a lag in motion compared to the \textit{High} velocity. This lag should be taken into account when analyzing and comparing the velocities.

\subsection*{Motion Uncertainty}
\begin{figure}[htbp]
    \centering
    \includegraphics[width=0.9 \textwidth]{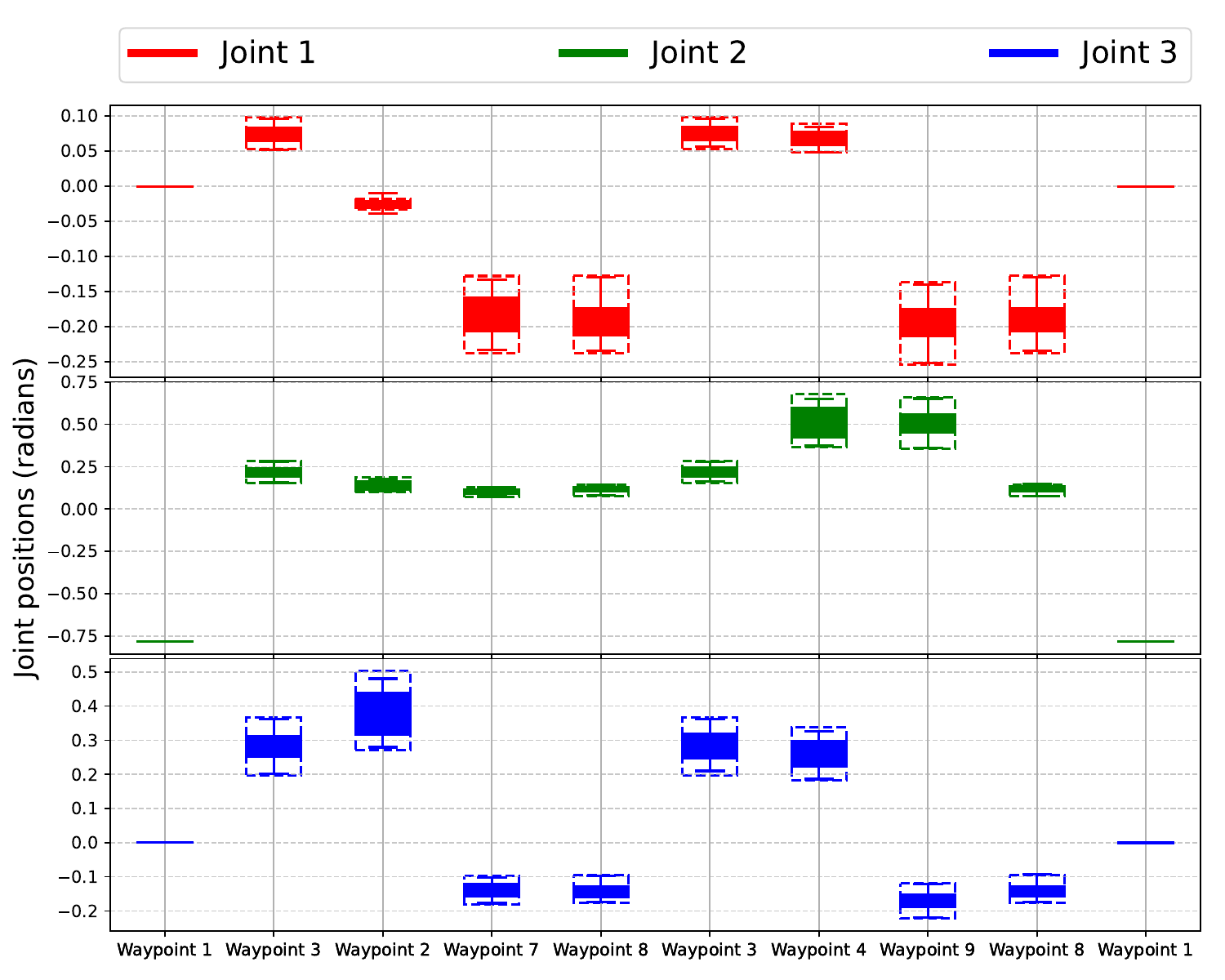}
    \caption{The sampled waypoints with added uncertainty are plotted for each waypoint, using all repetitions for $A=8$, $R=1$, and $V=\text{\textit{High}}$. We chose $A=8$ because it traverses all the waypoints multiple times, providing a representative example of the entire dataset. The sampled waypoints with uncertainty are visualized using a box plot, where the boxes represent the IQR and the whiskers extend to $1.5$ times the IQR. The maximum allowed uncertainty is indicated with dashed rectangles, showing that the noisy waypoints consistently fall within this range.}
    \label{fig:uncertainty_validation}
\end{figure}

As mentioned in the Methods section, we manually added multiplicative uniform noise $\epsilon \sim \mathcal{U}(0.7, 1.3)$ to the first three joints of the robot ($q_1, q_2, q_3$) for each waypoint in the sequence, except for waypoint 1. Figure~\ref{fig:uncertainty_validation} shows the actual samples for each waypoint of the first three joints. The filled boxes represent the interquartile range (IQR) of the noisy waypoints, while the whiskers extend to 1.5 times the IQR. The maximum range of uncertainty for each waypoint is illustrated using a dashed rectangle. It is evident that the noisy waypoints consistently fall within the range defined by the specified uncertainty.

\subsection*{CSI Synchronization}
To validate the synchronization of WiFi signals captured by the three sniffers, we analyzed the normalized cross-correlation of the WiFi CSI amplitude signals. This metric measures the similarity of the signal patterns and ensures that the data streams are temporally aligned across the devices.

\begin{table}[htbp]
\centering
\caption{CSI Normalized Cross-Correlation Statistics.}
\begin{tabular}{|c|c|c|c|c|c|}
\hline
Pair & Average  ($\pm$ STD) & Minimum  & Maximum & IQR  \\
\hline
CSI ($M=1\&2$) &  $9.31 \times 10^{-1} \pm  1.00  \times 10^{-1}$ & $8.74 \times 10^{-1}$ & $9.66 \times 10^{-1}$ & $1.45 \times 10^{-2}$ \\
CSI ($M=1\&3$) & $9.24 \times 10^{-1} \pm 9.00 \times 10^{-2}$ & $8.75 \times 10^{-1}$ & $9.61 \times 10^{-1}$ & $1.18 \times 10^{-2}$ \\
CSI ($M=2\&3$) & $9.23 \times 10^{-1} \pm 1.00 \times 10^{-1}$ & $8.84 \times 10^{-1}$ & $9.64 \times 10^{-1}$ & $1.44 \times 10^{-2}$ \\
\hline
\end{tabular}
\label{tab:csi_sync}
\end{table}

\begin{figure}[htbp]
    \centering
    \includegraphics[width=0.95 \textwidth]{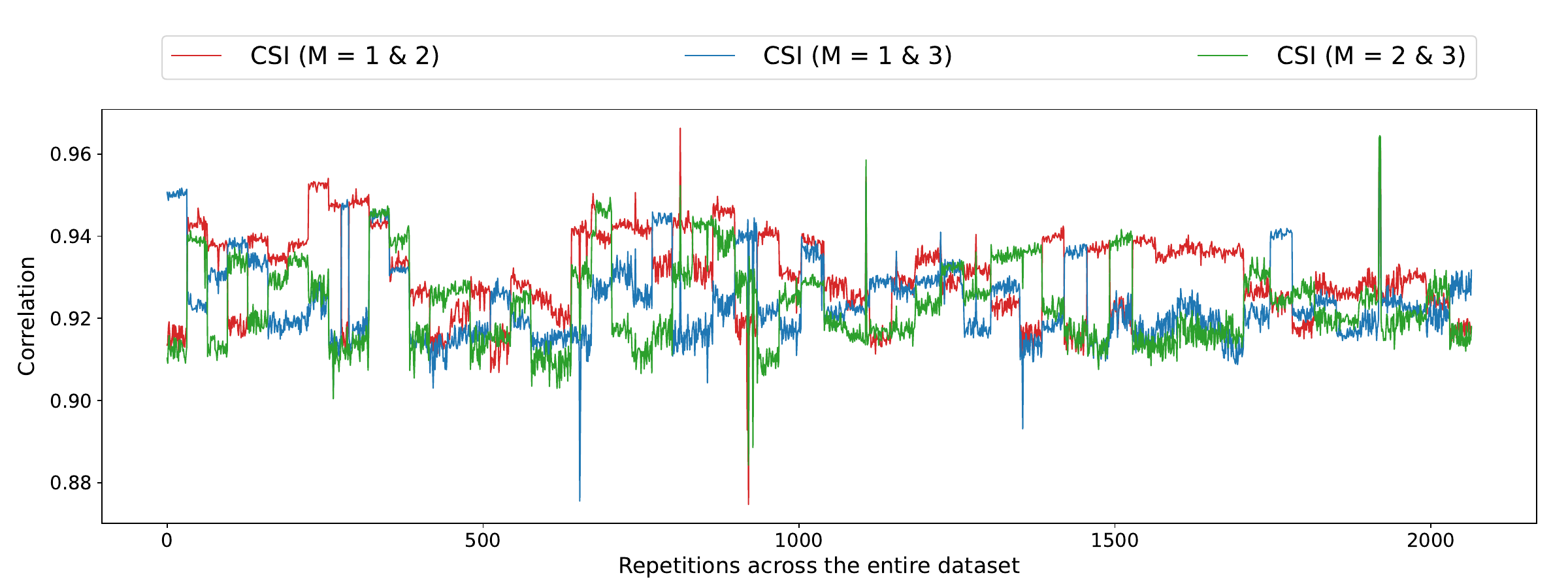}
    \caption{The normalized cross-correlation values for the WiFi CSI modality across all repetitions demonstrate strong synchronization among the three modules.}
    \label{fig:wifi_norm_corr}
\end{figure}

Figure \ref{fig:wifi_norm_corr} presents the normalized cross-correlation of each pair of sniffers for each repetition. Table \ref{tab:csi_sync} summarizes the statistics of the normalized cross-correlation values. As observed, the average normalized correlation values for all sniffer pairs exceed $9.23 \times 10^{-1}$, with a minimum value of $8.74 \times 10^{-1}$ and a maximum value of $9.66 \times 10^{-1}$. The IQR values are also small (e.g., $1.18 \times 10^{-2}$ to $1.45 \times 10^{-2}$), indicating low variability among the central data points. These results confirm that the CSI signals are highly linearly correlated, demonstrating strong synchronization across the sniffers.

\subsection*{Video Synchronization}
To validate the synchronization of the video recordings captured by the three cameras, we analyzed the normalized cross-correlation of pixel intensity values extracted from the video frames. These metrics measure the similarity of pixel patterns across the video streams, ensuring that the cameras are temporally aligned. The normalized cross-correlation values validate the synchronization, showing consistent similarity across the dataset. These findings demonstrate stable temporal alignment between the cameras, ensuring reliable synchronization.

\begin{figure}[htbp]
    \centering
    \includegraphics[width=0.95 \textwidth]{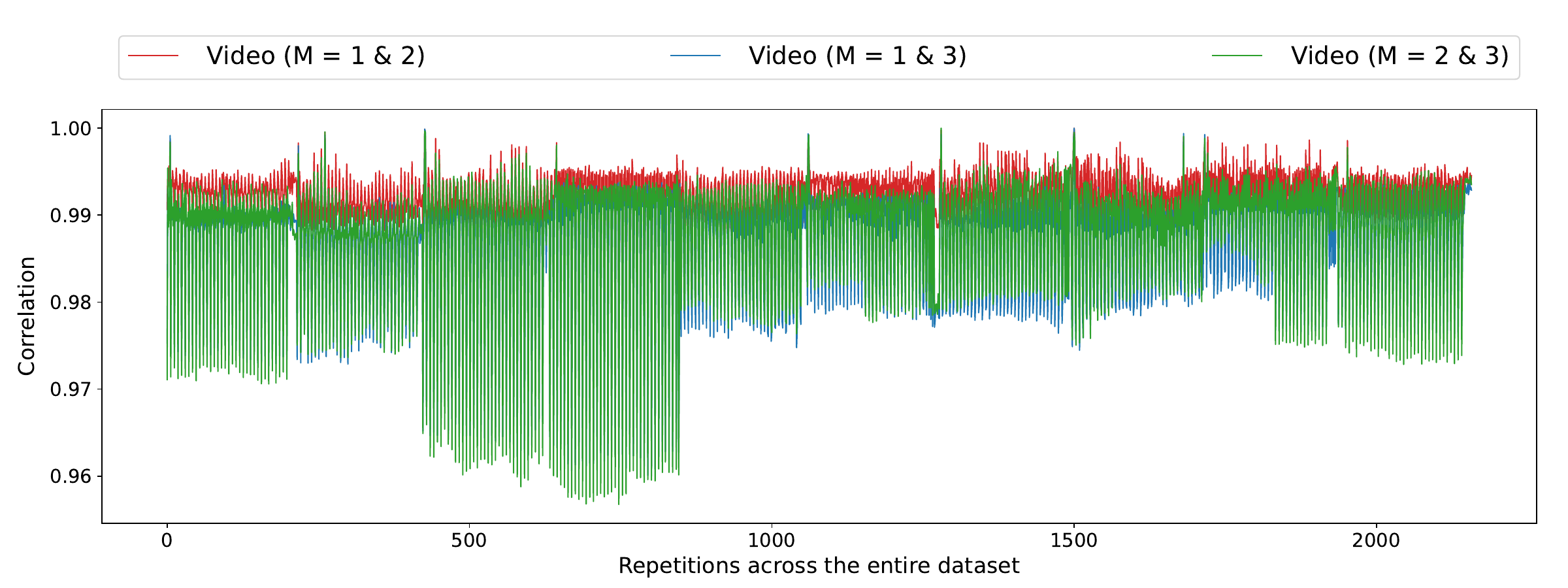}
    \caption{The normalized cross-correlation values for the video modality across all repetitions demonstrate strong synchronization among the three modules.}
    \label{fig:video_norm_corr}
\end{figure}

Figure \ref{fig:video_norm_corr} illustrates the normalized cross-correlation values across all repetitions. The x-axis denotes the repetitions, while the y-axis represents the magnitude of the correlation. These results validate the consistent synchronization of the video data across the three cameras, demonstrating strong alignment. Building on these observations, Table~\ref{tab:video_corr} presents key statistical metrics for the normalized cross-correlation values, including the mean, standard deviation, minimum, maximum, and IQR. These metrics quantitatively confirm the strong synchronization across all pairs with minimal variation.

\begin{table}[htbp]
\centering
\caption{Video Normalized Cross-Correlation Results.}
\begin{tabular}{|c|c|c|c|c|}
\hline
Pair & Average ($\pm$ STD) & Minimum & Maximum & IQR \\
\hline
Video ($M=1\&2$) & $9.92 \times 10^{-1} \pm 2.90 \times 10^{-3}$ & $9.81 \times 10^{-1}$ & $1.00$ & $4.12 \times 10^{-3}$ \\
Video ($M=1\&3$) & $9.87 \times 10^{-1} \pm 6.85 \times 10^{-3}$ & $9.60 \times 10^{-1}$ & $1.00$ & $7.81 \times 10^{-3}$ \\
Video ($M=2\&3$) & $9.88 \times 10^{-1} \pm 7.45 \times 10^{-3}$ & $9.57 \times 10^{-1}$ & $1.00$ & $5.32 \times 10^{-3}$ \\
\hline
\end{tabular}
\label{tab:video_corr}
\end{table}

\subsection*{Audio Synchronization}

To assess the synchronization of audio signals captured by the microphones, we computed the normalized cross-correlation of their spectrograms. This ensures that audio signals recorded by the devices are aligned in time, a key requirement for synchronization.

\begin{table}[htbp]
\centering
\caption{Audio Normalized Cross-Correlation Results.}
\begin{tabular}{|c|c|c|c|c|}
\hline
Pair & Average ($\pm$ STD)  & Minimum & Maximum & IQR\\
\hline
Audio ($M=1\&2$) & $7.85 \times 10^{-1} \pm  1.04  \times 10^{-2}$ & $7.77 \times 10^{-1}$ & $7.88 \times 10^{-1}$ & $1.38 \times 10^{-3}$\\
Audio ($M=1\&3$) &  $7.84 \times 10^{-1} \pm 7.36 \times 10^{-3}$ & $7.73 \times 10^{-1}$ & $7.86 \times 10^{-1}$ & $8.62 \times 10^{-4}$ \\
Audio ($M=2\&3$) & $7.85 \times 10^{-1} \pm 8.89 \times 10^{-3}$ & $7.76 \times 10^{-1}$ & $7.87 \times 10^{-1}$ & $1.21 \times 10^{-3}$\\
\hline
\end{tabular}
\label{tab:audio_sync}
\end{table}

\begin{figure}
    \centering
    \includegraphics[width=0.95 \textwidth]{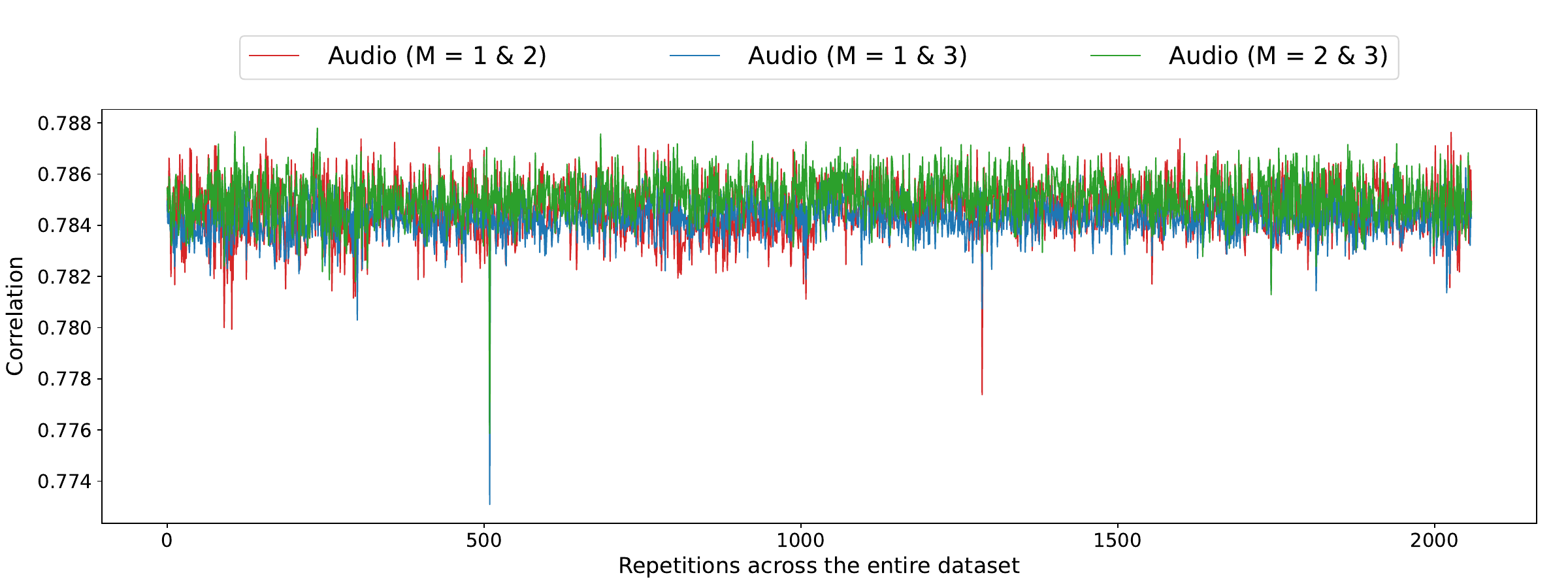}
    \caption{The normalized cross-correlation values for the audio modality across all repetitions demonstrate strong synchronization among the three modules.}
    \label{fig:audio_norm_corr}
\end{figure}

The normalized cross-correlation results confirm the synchronization of audio signals across the devices. The small standard deviations in the normalized values indicate stable synchronization with minimal deviations across the dataset. Figure \ref{fig:audio_norm_corr} illustrates the normalized cross-correlation for each pair of microphones across all repetitions, with each colored line representing a pair. Table \ref{tab:audio_sync} provides a detailed summary of the normalized cross-correlation statistics. As shown, the average normalized correlation values for all microphone pairs are above $7.84 \times 10^{-1}$, with minimum values ranging from $7.73 \times 10^{-1}$ to $7.77 \times 10^{-1}$ and maximum values close to $7.88 \times 10^{-1}$. With IQR values ranging from $8.62 \times 10^{-4}$ to $1.38 \times 10^{-3}$, the central data points show minimal variation. These results confirm that audio signals are highly synchronized between devices.

\begin{figure}[htbp]
    \centering
    \includegraphics[width=0.95 \textwidth]{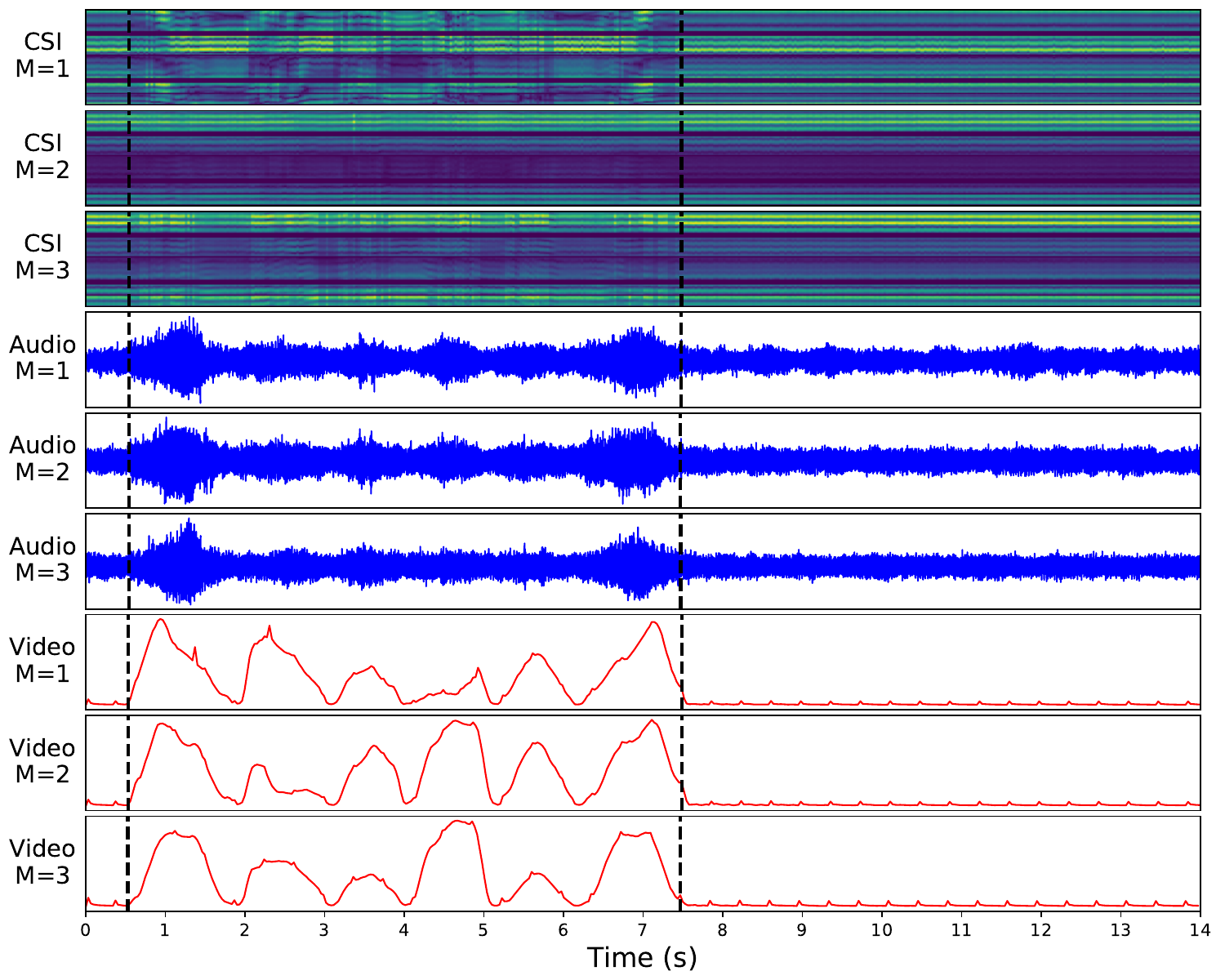}
    \caption{The plot shows the data from three sensor modules across CSI, audio, and video modalities. For CSI, a heatmap of subcarrier amplitudes is used; for audio, the waveform of the audio signal is presented, showing variations in amplitude over time; and for video, pixel-wise variance (squared differences) between consecutive frames is depicted. The start and end of the activity performed by the robot are indicated by black dashed lines. Comparing the signals from different sensors demonstrates their synchronization in time with respect to each other. This plot is taken from one of the repetitions (a.k.a. samples) in the dataset with $A = 0$, $R = 1$, and $V = \text{\textit{High}}$.}
    \label{fig:synchronization_validation_all}
\end{figure}

Additionally, Figure~\ref{fig:synchronization_validation_all} illustrates the data from various sensor modalities, showing the start and end times of the activity performed by the robot during one of the repetitions in the dataset. This figure also validates the synchronization between the different sensors and modules in the dataset.

\section*{Code availability}
The entire dataset is available for download from our Figshare repository \cite{figshare_dataset_2024}. The interested readers are encouraged to visit our GitHub repository (\href{https://github.com/SiamiLab/RoboMNIST}{https://github.com/SiamiLab/RoboMNIST}), where example Python notebooks for loading and visualizing our data are provided.

\begin{itemize}
    \item \texttt{wifi\_csi\_read.ipynb}: This Python notebook loads a CSI \textit{json} file from a repetition and visualizes the CSI amplitudes and RSS values.
    \item \texttt{true\_trajectory\_read.ipynb}: This Python notebook loads a true trajectory \textit{json} file from a repetition and visualizes the robot's motion.
\end{itemize}

The repository also includes the complete set of codes along with detailed explanations on how the robots were controlled, data was collected, and synchronization was achieved across different modules and sensors, and it provides instructions on how to reproduce these processes.

\section*{Acknowledgements}
This material is based upon work supported in part by grants ONR N00014-21-1-2431, NSF 2121121, the U.S. Department of Homeland Security under Grant Award Number 22STESE00001-01-00, and by the Army Research Laboratory under Cooperative Agreement Number W911NF-22-2-0001 (KB, RZ, EM, MS). The views and conclusions contained in this document are solely those of the authors and should not be interpreted as representing the official policies, either expressed or implied, of the U.S. Department of Homeland Security, the Army Research Office, or the U.S. Government.

\section*{Author Contributions}
Kian Behzad contributed significantly to hardware development, the experimental setup, and data collection, managed the GitHub repository for code availability and the Figshare repository for data sharing, and wrote the initial draft of the manuscript with feedback from all authors. Rojin Zandi led Wi-Fi and audio validation, contributed to data collection, and helped develop the experimental setup. Elaheh Motamedi performed video validation, contributed to data collection, and helped develop the experimental setup. Hojjat Salehinejad, as a senior author, contributed to shaping the research and provided critical feedback. Milad Siami conceived the project, secured funding, provided experimental resources, supervised all aspects, and guided the study throughout. All authors reviewed and approved the final manuscript.

\section*{Competing interests}
The authors declare no competing interests.


\bibliographystyle{unsrt}
\bibliography{mybibliography}

\end{document}